\documentclass[nohyperref]{article}

\def \dt {\mathcal{D}_{\mathrm{train}}}

\def \dout {\mathcal{D}_{\mathrm{out}}}
\def \dmix {\mathcal{D}_{\mathrm{mix}}}
\def \pt {P_{\mathrm{s}}}
\def \pte {P_{\mathrm{t}}}
\def \pout {P_{\mathrm{out}}}
\def \pmix {P_{\mathrm{mix}}}

\def \cX {\mathcal{X}}
\def \cY {\mathcal{Y}}

\def \vx {\boldsymbol{x}}

\usepackage{microtype}
\usepackage{graphicx}
\usepackage{subcaption}
\usepackage{booktabs} 

\usepackage{hyperref}

\usepackage[accepted]{icml2022}

\usepackage{amsmath}
\usepackage{amssymb}
\usepackage{mathtools}
\usepackage{amsthm}
\usepackage{subcaption}
\usepackage{booktabs}
\usepackage{epsfig}
\usepackage{setspace}
\usepackage{comment}
\usepackage{xcolor}
\usepackage{multirow}
\usepackage[capitalize,noabbrev]{cleveref}

\theoremstyle{plain}
\newtheorem{theorem}{Theorem}[section]
\newtheorem{proposition}[theorem]{Proposition}

\newtheorem{definition}[theorem]{Definition}

\theoremstyle{remark}

\usepackage[textsize=tiny]{todonotes}
\usepackage{colortbl}

\newcommand{\CC}[1]{\cellcolor[rgb]{.8, .8, .9}}

\usepackage[textsize=tiny]{todonotes}

\icmltitlerunning{Exploring Out-of-Distribution Data for Re-balancing Long-tailed Datasets}

\begin{document}

\twocolumn[
\icmltitle{Open-Sampling: Exploring Out-of-Distribution Data for Re-balancing Long-tailed Datasets}

\icmlsetsymbol{equal}{*}

\begin{icmlauthorlist}
\icmlauthor{Hongxin Wei}{ntu}
\icmlauthor{Lue Tao}{nuaa}
\icmlauthor{Renchunzi Xie}{ntu}
\icmlauthor{Lei Feng}{cqu}
\icmlauthor{Bo An}{ntu}
\end{icmlauthorlist}

\icmlaffiliation{ntu}{Nanyang Technological University, Singapore}
\icmlaffiliation{nuaa}{Nanjing University of Aeronautics and Astronautics, Nanjing, Jiangsu, China}
\icmlaffiliation{cqu}{Chongqing University, Chongqing, China}

\icmlcorrespondingauthor{Renchunzi Xie}{XIER0002@e.ntu.edu.sg}

\icmlkeywords{long-tailed, out-of-distribution, }

\vskip 0.3in
]

\printAffiliationsAndNotice{\icmlEqualContribution} 
\begin{abstract}
Deep neural networks usually perform poorly when the training dataset suffers from extreme \textit{class imbalance}. 
Recent studies found that directly training with out-of-distribution data (i.e., open-set samples) in a semi-supervised manner would harm the generalization performance.
In this work, we theoretically show that out-of-distribution data can still be leveraged to augment the minority classes from a Bayesian perspective. Based on this motivation, we propose a novel method called \textit{Open-sampling}, which utilizes open-set noisy labels to re-balance the class priors of the training dataset. For each open-set instance, the label is sampled from our pre-defined distribution that is complementary to the distribution of original class priors.
We empirically show that Open-sampling not only re-balances the class priors but also encourages the neural network to learn separable representations.
Extensive experiments demonstrate that our proposed method significantly outperforms existing data re-balancing methods and can boost the performance of existing state-of-the-art methods.
\end{abstract}

\section{Introduction}

The success of deep neural networks (DNNs) heavily relies on large-scale datasets with balanced distribution \cite{krizhevsky2009learning, ILSVRC15}. However, in real-world applications like autonomous driving and medical diagnosis, large-scale datasets naturally exhibit imbalanced and long-tailed distributions, i.e., a few classes (majority classes) occupy most of the data while most classes (minority classes) are under-represented \cite{zhou2017places, van2018inaturalist, lin2014microsoft}. It has been shown that training on long-tailed datasets leads to poor generalization performance, especially on the minority classes \cite{zhou2020BBN,liu2019large,kang2019decoupling}. Thus, designing effective algorithms to handle class imbalance is of great practical importance.

In the literature, a popular direction in long-tailed learning is to re-balance the data distribution by data re-sampling \cite{japkowicz2002class, he2009learning}.
For example, Over-sampling \cite{byrd2019effect, shen2016relay} repeats samples from under-presented classes, but it usually causes over-fitting to the minority classes.
To alleviate the over-fitting issue, synthesized novel samples are introduced to augment the minority classes without repetition \cite{chawla2002smote}. As a result, the model is still error-prone due to noise in the synthesized samples \cite{cui2019class}
A recent work \cite{yang2020rethinking} introduced unlabeled-in-distribution data to compensate for the lack of training samples, and showed that directly adding unlabeled data from mismatched classes (\emph{i.e.}, out-of-distribution data) by semi-supervised learning hurts the generalization performance. 
These data augmentation methods normally require in-distribution data with precise labels for selected classes. However, such kind of data would be extremely hard to collect in real-world scenarios, due to the expensive labeling cost. This fatal weakness of previous methods motivates us to explore the possibility of using \textit{out-of-distribution} (OOD) data for long-tailed imbalanced learning.

In this paper, we theoretically show that out-of-distribution data (i.e., open-set samples) could be leveraged to augment the minority classes from a Bayesian perspective. 
Based on this motivation, we propose a simple yet effective method called Open-sampling, which uses open-set noisy labels to re-balance the label priors of the training dataset. For each OOD instance, the label is sampled from our pre-defined distribution that is complementary to the original class priors. To alleviate the over-fitting issue on the minority classes, a class-dependent weight is used in the training objective to provide stronger regularization on the minority classes than the majority classes. In this way, the open-set noisy labels could be used to re-balance the class priors while retaining their non-toxicity.

To provide a comprehensive understanding, we conduct a series of analyses to illustrate the properties of the proposed Open-sampling method. From these empirical analyses, we show that: 1) the Complementary Distribution is superior to the commonly used Class Balanced distribution (CB) as the former is closer to the uniform distribution, which reduces the harmfulness of the open-set noisy labels; 
2) real-world datasets with large sample size are the best choices for the open-set auxiliary dataset in Open-sampling and the diversity (i.e., number of classes) is not an important factor in the method;
3) the Open-sampling method not only re-balances the class prior, but also promotes the neural network to learn more separable representations.

To the best of our knowledge, we are the first to explore the benefits of OOD instances in learning from long-tailed datasets. To verify the effectiveness of our method, we conduct experiments on four long-tailed image classification benchmark datasets, including long-tailed CIFAR10/100 \cite{krizhevsky2009learning}, CelebA-5 \cite{liu2015faceattributes, kim2020m2m}, and Places-LT \cite{zhou2017places}. Empirical results show that our method can be easily incorporated into existing state-of-the art methods to enhance their performance on long-tailed imbalanced classification tasks. Furthermore, experimental results validate that our method could also achieve impressive performance for detecting OOD examples under class-imbalanced setting. Code and data are publicly available at  \url{https://github.com/hongxin001/open-sampling}.

\section{Imbalanced Learning with OOD instances}
\label{sec:method}

\subsection{Background}

In this work, we consider a multi-class classification problem, where the input space is denoted by $\cX \in \mathbb{R}^{d}$ and the label space $\cY$ is $\{1, \ldots, K\}$. We denote by $\dt = \{ (\vx_i, y_i) \}^N_{i=1} \in \cX \times \cY$ the training dataset with $N$ samples. Let $n_j$ be the number of samples in class $j$, then $N=\sum^{K}_{j=1}n_j$.
Let $\pt(X, Y)$ define the underlying training (source) distribution and $\pte(X, Y)$ define the test (target) distribution. Generally, the class imbalance problem assumes that the test data has the same class conditional probability as the training data, i.e., $\pt(X|Y)=\pte(X|Y)$, while their class priors are different, i.e., $\pt(Y) \neq \pte(Y)$. 

Besides, we consider an unlabelled auxiliary dataset $\dout^{(\vx)}=\{\Tilde{\vx}_i\}^{M}_{i=1} \in \cX$ consisting of $M$ open-set instances, and we have $M \gg N$. These open-set instances are also known as OOD data as they are sampled from $\pout(X)$, which is disjoint from $\pt(X)$. In real-world scenarios, it is easy to obtain such auxiliary datasets, which are commonly used in OOD detection tasks. In what follows, we may assign each open-set instance $\Tilde{\vx}_i$ with a random label $\Tilde{y}_i \in \cY$, which is independently sampled from an appropriate label distribution $\pout(Y)$ over $\cY$. We denote by $\dout=\{(\Tilde{\vx}_i, y_i)\}^{M}_{i=1}$ the auxiliary dataset with randomly sampled noisy labels.

\subsection{Theoretical Motivation}

From a Bayesian perspective, the prediction of a Bayes classifier is generally made as follows:
\begin{equation}
\label{eq:bayesian}
y^{*} = \underset{y\in\cY}{\arg\max} ~P(y|\vx) = \underset{y\in\cY}{\arg\max} ~P(\vx|y)P(y),
\end{equation}
where $P(\vx)$ and $P(y)$ represent $P(X=\vx)$ and $P(Y=y)$, respectively.
Unfortunately, the predicted posterior probability in Eq.~(\ref{eq:bayesian}) becomes unreliable when there is a large discrepancy between the class priors of training and test distributions. Specifically, the class prior of the test distribution is usually balanced (i.e., a uniform distribution over labels), while the training dataset exhibits a long-tailed class distribution. Formally, we have $\pt(Y=i) \neq \pte(Y=i)$ for any class $i \in \cY$, and $\pt(Y=j) \ll \pte(Y=j)$ for a minority class $j \in \cY$. 

To re-balance the class priors of the training dataset, existing re-sampling methods mostly augment the minority classes with extra \textit{in-distribution} samples with precise labels, e.g., duplicated examples \cite{buda2018systematic}, synthetic generation \cite{he2008adasyn}, and interpolation \cite{chawla2002smote}. However, due to their strict in-distribution constraints on data distribution and label quality, generating and collecting those samples are challenging and expensive, especially for minority classes. In this paper, we instead try to break through the constraints by demonstrating that OOD instances with noisy labels are actually useful for re-balancing the training dataset.

We start by presenting an intriguing fact in the following theorem (proof in Appendix~\ref{app:proofs_1}), which demonstrates that augmenting the training data with open-set instances and uniformly random labels can be non-toxic. 


\begin{theorem}
\label{thm:bayes_fix}
Assume that $\pout(Y)$ is the discrete uniform distribution over the label space $\cY$. Let the augmented dataset be $\dmix = \dt \cup \dout$, and $\pmix(X, Y)$ be the underlying data distribution of $\dmix$, then we have
$$
\underset{y\in\cY}{\arg\max}~\pmix(\vx|y)\pmix(y) = \underset{y\in\cY}{\arg\max}~\pt(\vx|y)\pt(y).
$$
\end{theorem}

Theorem \ref{thm:bayes_fix} indicates that, when the labels are uniformly sampled from the in-distribution label space, the prediction of the Bayes classifier is unchanged after augmenting the training dataset with the OOD instances\footnote{The discovery of the non-toxicity of OOD instances is not new and has been studied in ODNL \cite{wei2021open}. Our unique contribution here is to theoretically prove this property from a Bayesian perspective. We provide a detailed comparison for these two works in Subsection \ref{sec:sampling}. }. In this way, the theoretical result motivates us to exploit the potential value of OOD instances to repair class imbalance.

\begin{figure}[h]
    \centering
    \includegraphics[width=0.30\textwidth]{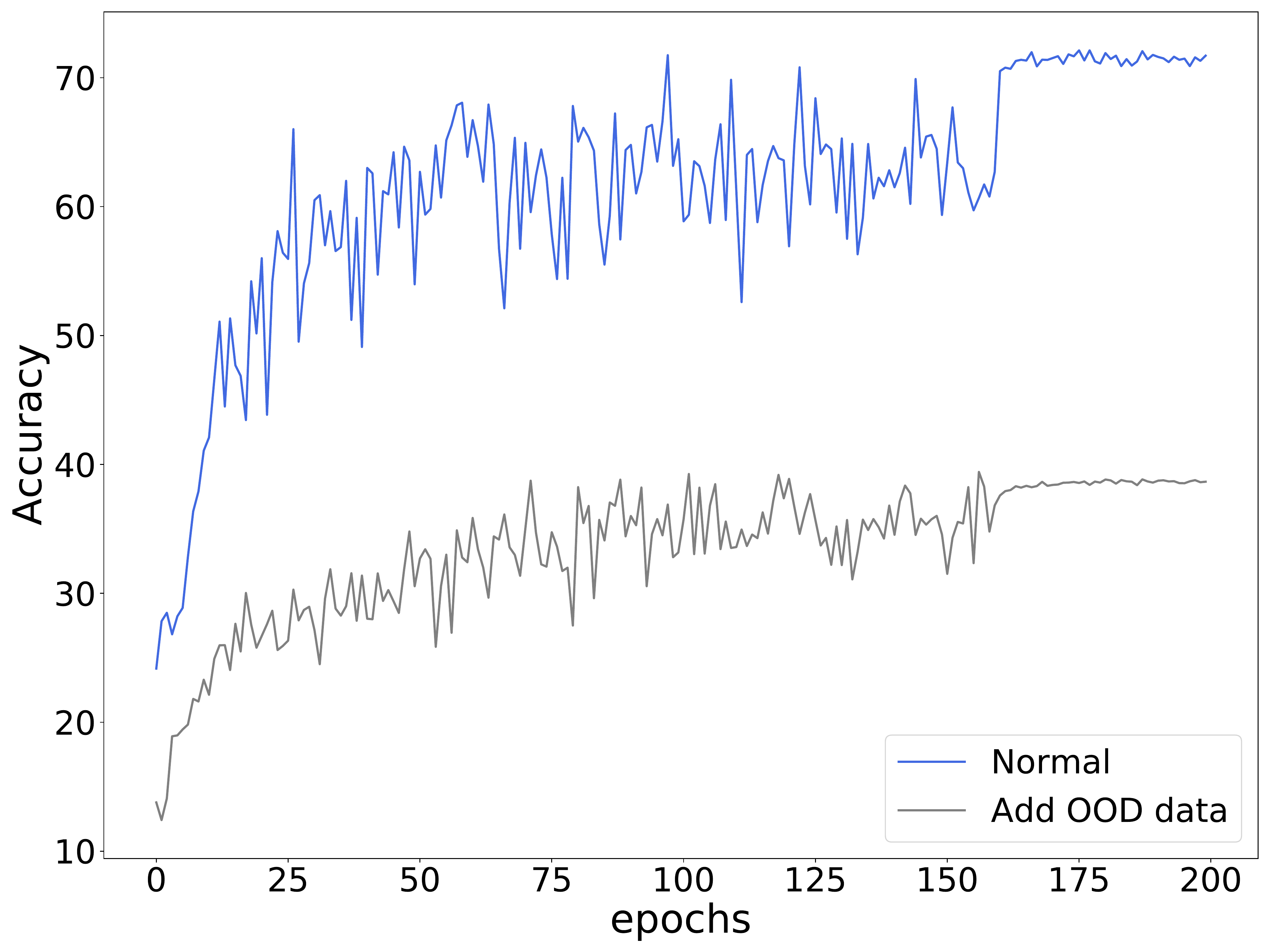}
    \vspace{-0.3cm}
    \caption{The test accuracy of models under different training epochs. Blue color denotes the model trained on long-tailed CIFAR-10 with ResNet-34, and gray color denotes the model trained on long-tailed CIFAR-10 and OOD data. We use instances from 300K Random Images \cite{hendrycks2019oe} as OOD data and label as a minority class---``9''. The comparison implies that simply adding OOD data into training may downgrade the generalization performance.}
    \label{fig:example_ood}
    \vspace{-5pt}
\end{figure}

Although using the uniform distribution as $\pout(Y)$ will not downgrade the Bayes classifier as shown above, the resulting classifier is not yet optimal, and its corresponding partition is still far away from the optimal decision boundary on the test distribution, since the label distribution $\pmix(Y)$ still remains largely imbalanced. 
In the following, we will show that a better classifier can be obtained by exploring the trade-off between re-balancing the label distribution and keeping non-toxicity.

\subsection{Open-sampling}
\label{sec:sampling}
Motivated by the previous analysis, we propose to exploit the open-set auxiliary dataset to improve the generalization under class-imbalanced settings. With a proper label distribution, OOD instances with dynamic labels can be used to re-balance the class priors while retaining their non-toxicity. We start by giving the following definition.

\begin{definition}[Complementary Distribution]
Complementary Distribution (CD) is a label distribution for the auxiliary dataset to re-balance the class priors of the original dataset. In particular, Minimum Complementary Distribution (MCD) is the complementary distribution that requires the smallest number of auxiliary instances to re-balance the original training set.
\end{definition}

Designing a proper Complementary Distribution to mitigate the class imbalance problem is a difficult problem that depends on the trade-off between re-balancing the class priors and keeping the non-toxicity of the added noisy labels. Intuitively, to re-balance the class priors, more OOD instances should be allocated into the minority classes than the majority classes. Meanwhile, the unequal number of OOD instances in different classes may shift the Bayes classifier. 

The above conflict can be understood naturally through a simple case. We consider a binary classification task with $K=2$, and let the sample numbers of the two classes be $n_1$ and $n_2$, respectively. Without loss of generality, let $n_1 > n_2$. It is straightforward to verify that one of the optimal allocation to re-balance the class priors is simply appending $n_1 - n_2$ extra OOD instances into the minority class. However, in such a way, all OOD instances are assigned the same label, thereby downgrading the resulting classifier. 

To provide a straightforward view, we show in Figure~\ref{fig:example_ood} the performance of model trained with the extra OOD data. Indeed, in the extreme case where all OOD instances are assigned the same label, their toxicity would be enlarged during training, leading to poor generalization performance.

\textbf{Complementary Sampling Rate.} To find a ``sweet spot'', we propose the following sampling rate that allows us to achieve a smooth transition from the MCD to the uniform distribution. We denote CD as $\Gamma$, MCD as $\Gamma^m$, and the complementary sampling rate for class $j$ as $\Gamma_j$, then we have the following proposition.

\begin{proposition} [Complementary Sampling Rate]
\label{prop:complementary} $\Gamma_j = (\alpha-\beta_j)/(K \cdot \alpha-1)$, where $\beta_j = \frac{n_j}{\sum^{K}_{i=1} n_i}$. Then, (i) $\sum^{K}_{i=1} \Gamma_i = 1$; (ii) $\Gamma = \Gamma^m$ if $\alpha = \max_j (\beta_j)$; (iii) $\Gamma_j \to 1/K$ as $\alpha \to \infty$.  
\end{proposition}

The hyperparameter $\alpha \in \mathbb{R}^{+} \geq \max_j (\beta_j)$ controls the trade-off bewteen the MCD and uniform distribution. As shown in Proposition \ref{prop:complementary}, when $\alpha =\max_j (\beta_j)$, it recovers the MCD. With a larger value for the $\alpha$, the label distribution of the auxiliary dataset would be closer to a uniform distribution. The proof of Proposition \ref{prop:complementary} is provided in Appendix~\ref{app:proofs_2}.

\begin{figure}
     \centering
     \begin{subfigure}[b]{0.15\textwidth}
         \centering
         \includegraphics[width=\textwidth]{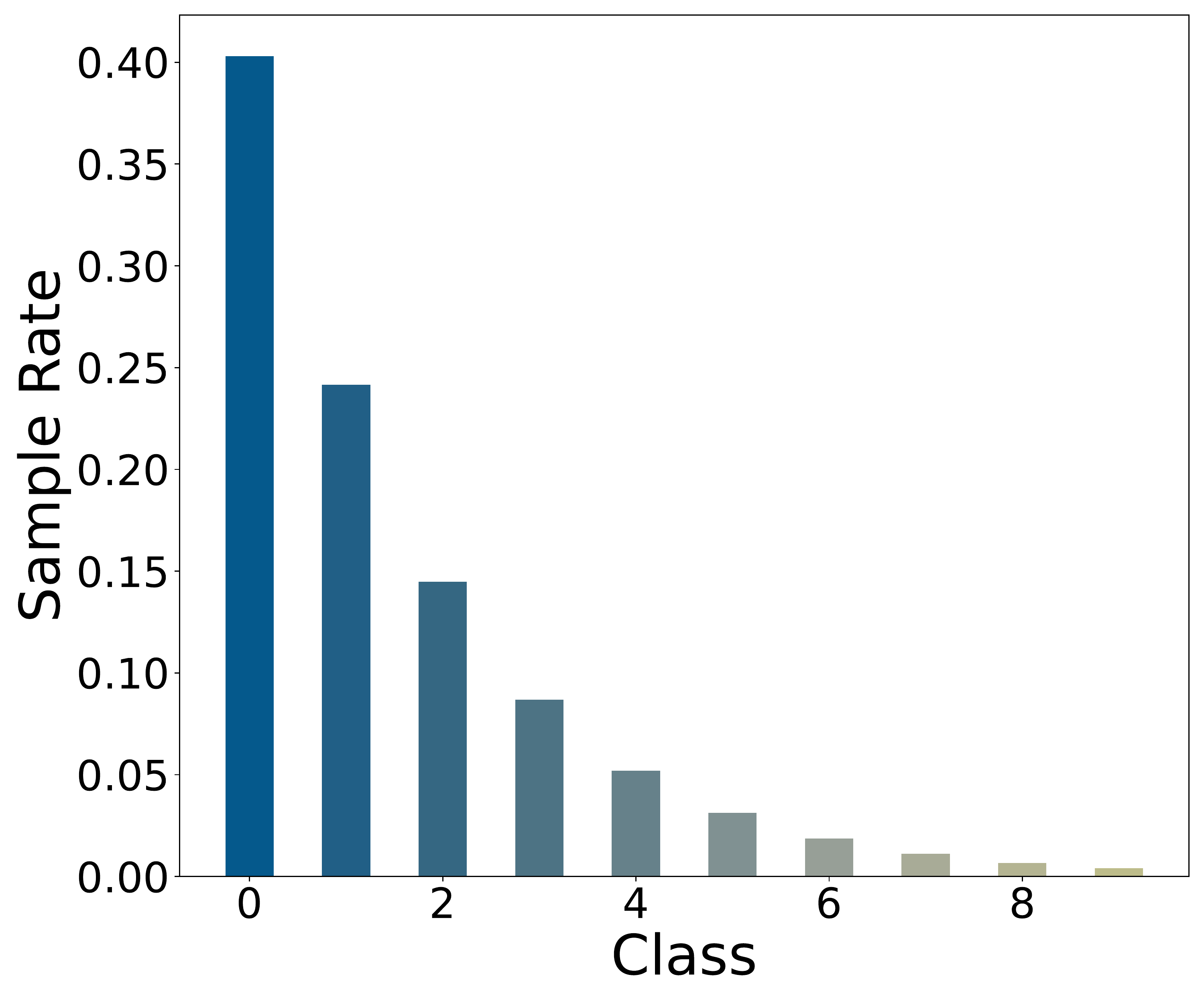}
         \caption{Original}
         \label{fig:origin}
     \end{subfigure}
     \hfill
     \begin{subfigure}[b]{0.15\textwidth}
         \centering
         \includegraphics[width=\textwidth]{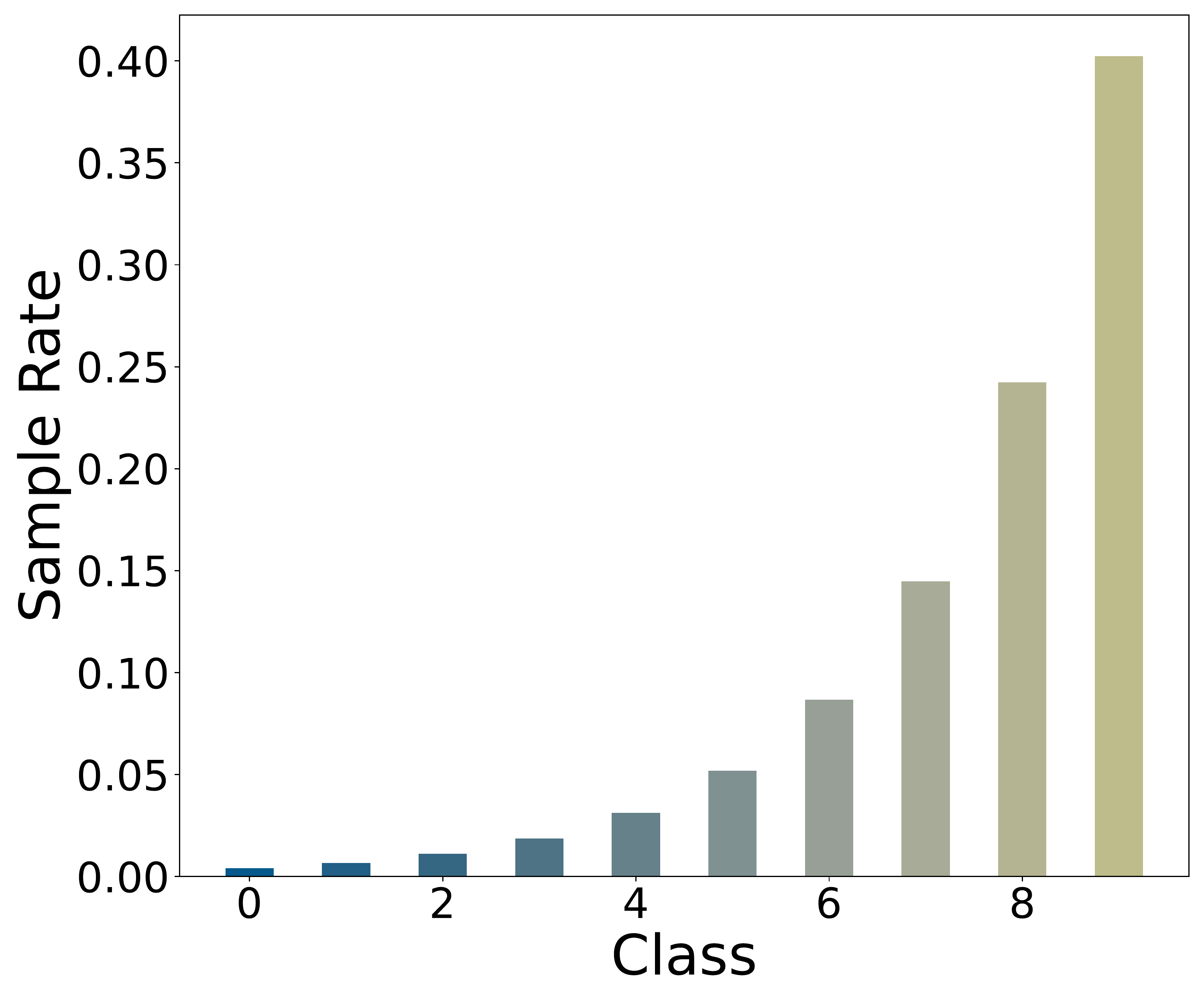}
         \caption{CB}
         \label{fig:cb}
     \end{subfigure}
     \hfill
     \begin{subfigure}[b]{0.15\textwidth}
         \centering
         \includegraphics[width=\textwidth]{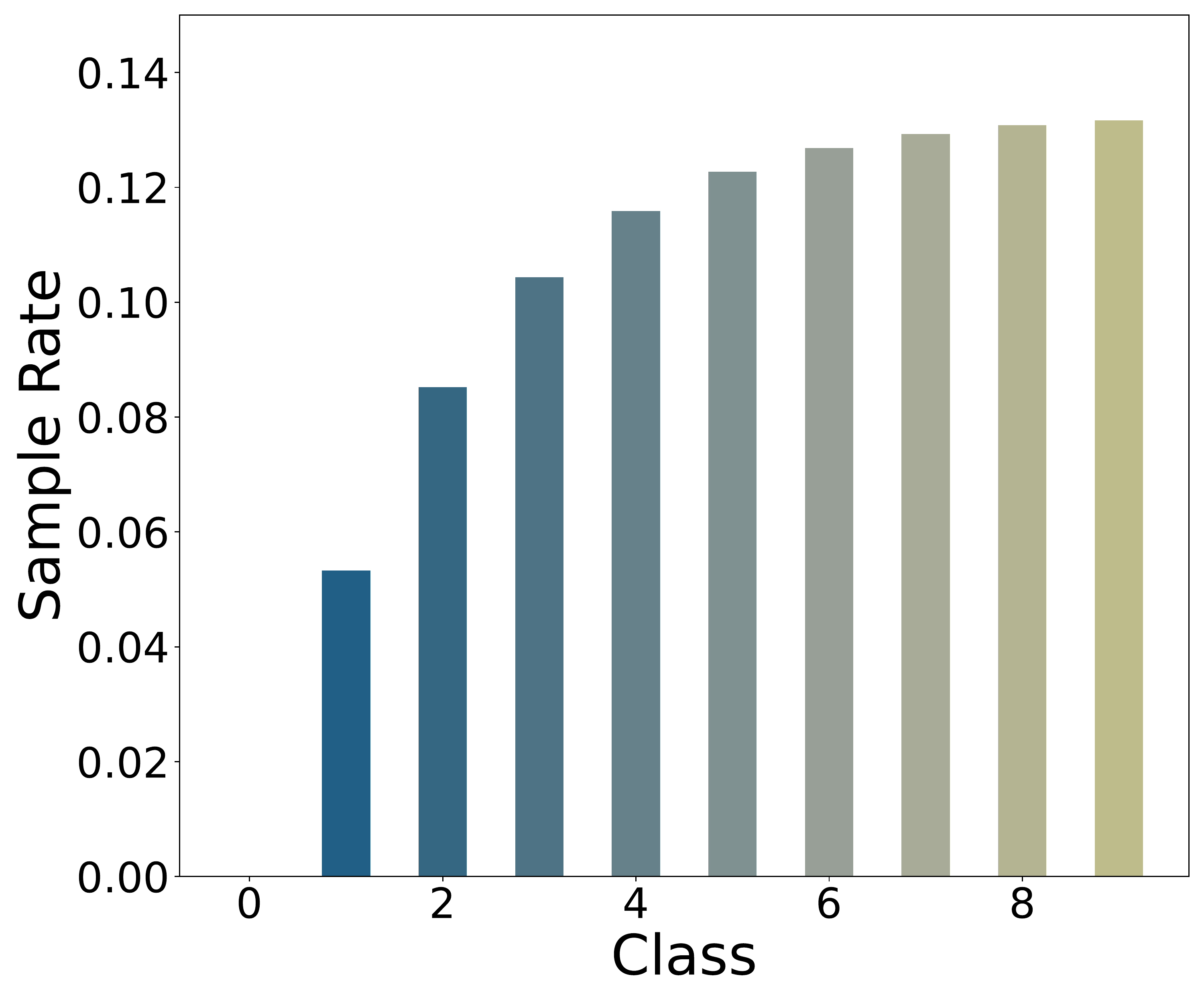}
         \caption{MCD}
         \label{fig:mcd}
     \end{subfigure}
     \caption{An illustration of label distributions for long-tailed CIFAR-10 dataset with imbalance ratio 100. (\ref{fig:origin}) Original label prior, (\ref{fig:cb}) Class Balanced distribution, (\ref{fig:mcd}) Minimum Complementary Distribution. }
    \label{fig:label_distribution}
    \vspace{-10pt}
\end{figure}

As shown in Figure~\ref{fig:label_distribution}, both the CB and MCD distributions exhibit an inverse relationship with the original label prior, i.e., the minority classes possess larger sampling rate than the majority classes. Compared with the CB distribution, our MCD distribution is flatter so that the instances would not be concentrated in a class. In such a manner, the harmfulness of open-set noisy labels would be alleviated to a large extent. The advantage of the MCD distribution is empirically supported in Section~\ref{sec:analyze}. 

With the Complementary Sampling Rate, we can then build an auxiliary dataset with OOD instances to re-balance the training dataset while retaining their non-toxicity.

\textbf{Training Objective}. To involve the sampled open-set noisy labels into the training, a natural idea is directly combining them with the original training dataset, and using the standard cross entropy as the training objective function:
\begin{align*}
\label{eq:loss}
\mathcal{L} &= \mathbb{E}_{\pmix(X, Y)}[\ell (f(\boldsymbol{x}; \boldsymbol{\theta}, y)] \\
&= \mathbb{E}_{\pmix(X, Y)}[- \mathbf{e}^{y} \log f(\boldsymbol{x}; \boldsymbol{\theta})],
\end{align*}
where 
$\boldsymbol{e}^{y} \in\{0,1\}^{K}$ denotes the one-hot vector whose $y$-th entry of $\boldsymbol{e}^{y}$ is 1. In each epoch, the labels of samples from the auxiliary dataset could be updated following the Complementary Distribution $\Gamma$.

However, the naive combination would consume too much capacity of the network on fitting the open-set noisy labels, making it hard to converge, especially when the sample size of the auxiliary dataset is much larger than that of the original training dataset. To handle this issue, we propose to use the loss on the auxiliary dataset as a regularization term as shown below:
\begin{equation*}
    \mathcal{L}_{\mathrm{reg}} = \mathbb{E}_{\widetilde{\boldsymbol{x}}\sim\pout{(X)}}\left[\ell\left(f(\widetilde{\boldsymbol{x}} ; \boldsymbol{\theta}), \widetilde{y}\right)\right],
\end{equation*}
where $\widetilde{y}$ is drawn from the Complementary Distribution $\Gamma$.

To further alleviate the over-fitting issue on the minority classes without sacrificing the performance on the majority classes, we explicitly introduce a class-dependent weighting factor $\omega_j$ based on the pre-defined Complementary Distribution. To make the total loss roughly in the same scale after applying $\omega_j$, we normalize $\omega$ so that $\sum^{K}_{j=1} \omega_j = K$. Then, the regularization item becomes:
\begin{equation*}
    \mathcal{L}_{\mathrm{reg}} = \mathbb{E}_{\widetilde{\boldsymbol{x}}\sim\pout{(X)}}\left[\omega_{\widetilde{y}} \cdot \ell\left(f(\widetilde{\boldsymbol{x}} ; \boldsymbol{\theta}), \widetilde{y}\right)\right],
\end{equation*}
where $\widetilde{y} \sim \Gamma$ and $\omega_{\widetilde{y}} = \Gamma_{\widetilde{y}} \cdot K$.
Now, the final training objective function is given as follows:
\begin{align}
\begin{split}
    \label{eq:final_loss}
    \mathcal{L}_{\mathrm{total}} &= \mathbb{E}_{((\boldsymbol{x},y)\sim\pt{(X,Y)})}\left[\ell\left(f(\boldsymbol{x} ; \boldsymbol{\theta}), y\right)\right] \\
    &+ \eta \cdot \mathbb{E}_{(\widetilde{\boldsymbol{x}})\sim\pout{(X)}}\left[\omega_{\widetilde{y}} \cdot \ell\left(f(\widetilde{\boldsymbol{x}} ; \boldsymbol{\theta}), \widetilde{y}\right)\right],
\end{split}
\end{align}
where $\eta$ controls the strength of the regularization term. The corresponding algorithm is provided in Appendix \ref{app:alg}.

As a data re-balancing technique, Open-sampling is orthogonal to existing methods, including the training objective based methods (e.g., LDAM \cite{cao2019learning}, Balanced Softmax \cite{Ren2020balms}), Decoupled training methods \cite{kang2019decoupling}, and self-supervised pre-trained methods \cite{yang2020rethinking}. Our method can be easily incorporated into these algorithms to further improve their generalization performance.
Given the original learning objective $\mathcal{L}_{\mathrm{imb}}$ of the existing methods, we can formalize the final objective as:
\begin{align}
    \mathcal{L}_{\mathrm{total}} &= \mathcal{L}_{\mathrm{imb}} + \eta \cdot \mathcal{L}_{\mathrm{reg}}.
\end{align}

\textbf{Relation to ODNL.} Recent work \cite{wei2021open} shows that open-set noisy labels could be applied to enhance the robustness against inherent noisy labels, which has some conceptual similarities to the proposed method in this work. Here, we summarize the main differences between ODNL \cite{wei2021open} and our work.

\begin{enumerate}

\item Problem setting: ODNL focuses on improving the robustness against noisy labels while our work considers the problem of learning from long-tailed imbalanced datasets. 

\item Technique: we proposed to sample the labels of OOD instances from the Complementary Distribution and add a weight factor to their losses, while they treats all the OOD instances equally by simply using a uniform distribution. In particular, ODNL can be seen as a special case of our method with a large value of $\alpha$. As analyzed in Figures \ref{fig:distribution}, \ref{fig:test_error_class}, and \ref{fig:alpha}, our Open-sampling consistently outperforms the variant with a uniform distribution or a large value of $\alpha$, which demonstrates the advantage of the proposed method.

\item Insight: ODNL aims to consume the extra representation capacity of neural networks to prevent over-fitting inherent noisy labels and show that their method helps the network converge to a flat minimum as SGD noises. In our work, OOD instances are applied to re-balance the label prior of the training dataset and the proposed method are shown to encourage the network to learn more separable representations. 
\end{enumerate}

\section{Experiments}
\label{sec:exp}
In this section, we evaluate our proposed method on long-tailed image classification datasets, including long-tailed CIFAR10/100 \cite{krizhevsky2009learning}, CelebA-5 \cite{liu2015faceattributes, kim2020m2m}, Places-LT \cite{zhou2017places}. Then we analyze the impact of $\eta$ by sensitivity analysis. The ablation study for the class-dependent weight is provided in Appendix \ref{app:results}. Finally, we conduct experiments to evaluate the performance of Open-sampling on OOD detection task under class-imbalanced setting. The datasets and implementation details are introduced in Appendix \ref{app:settings}.

\begin{table*}[!t]
\centering
\caption{Test accuracy (\%) of ResNet-32 on long-tailed CIFAR-10 and CIFAR-100 with various imbalance ratios. ``$\dagger$'' indicates the reported results from \cite{kim2020m2m}. The bold indicates the improved results by integrating our regularization. }
\label{tab:cifar10_results}
\renewcommand\arraystretch{1.0}
\resizebox{1.00\textwidth}{!}{
\setlength{\tabcolsep}{5mm}{
\begin{tabular}{c|ccc|ccc}
\toprule
Dataset & \multicolumn{3}{c|}{Long-tailed CIFAR-10} & \multicolumn{3}{c}{Long-tailed CIFAR-100} \\
\midrule
Imbalance Ratio & 100 & 50 & 10 & 100 & 50 & 10  \\
\midrule
Standard& 71.61 $\pm$ 0.21 & 77.30 $\pm$ 0.13 & 86.74 $\pm$ 0.41 & 37.59 $\pm$ 0.19 & 43.20 $\pm$ 0.30 & 56.44 $\pm$ 0.12 \\
SMOTE $^{\dagger}$& 71.50 $\pm$ 0.57 & - & 85.70 $\pm$ 0.25 & 34.00 $\pm$ 0.33 & - & 53.80 $\pm$ 0.93\\
CB-RW& 72.57 $\pm$ 1.30 & 78.19 $\pm$ 1.79 & 87.18 $\pm$ 0.95 & 38.11 $\pm$ 0.78 & 43.26 $\pm$ 0.87 & 56.40 $\pm$ 0.40 \\
CB-Focal& 70.91 $\pm$ 0.39 &  77.71 $\pm$ 0.57 & 86.89 $\pm$ 0.21 & 37.84 $\pm$ 0.80 & 42.96 $\pm$ 0.77& 56.09 $\pm$ 0.15\\
\textbf{Ours} & \textbf{77.62 $\pm$ 0.28} & \textbf{81.76 $\pm$ 0.51} & \textbf{89.38 $\pm$ 0.46} & \textbf{40.26 $\pm$ 0.65} & \textbf{44.77 $\pm$ 0.25} & \textbf{58.09 $\pm$ 0.29}\\
\midrule
LDAM-RW & 74.21 $\pm$ 0.61 & 78.86 $\pm$ 0.65 & 86.44 $\pm$ 0.78 & 29.02 $\pm$ 0.34 & 36.41 $\pm$ 0.84 & 54.23 $\pm$ 0.72\\
\textbf{+ Ours} &\textbf{75.19 $\pm$ 0.34} & \textbf{79.76 $\pm$ 0.44} & \textbf{87.28 $\pm$ 0.61} & \textbf{35.85 $\pm$ 0.62} & \textbf{42.18 $\pm$ 0.82} & \textbf{55.48 $\pm$ 0.59}\\
\midrule
LDAM-DRW & 78.08 $\pm$ 0.38& 81.88 $\pm$ 0.44& 87.49 $\pm$ 0.18 & 42.84 $\pm$ 0.25& 47.13 $\pm$ 0.28 & 57.18 $\pm$ 0.47\\
\textbf{+ Ours} & \textbf{79.82 $\pm$ 0.31} & \textbf{82.22 $\pm$ 0.45} & \textbf{87.83 $\pm$ 0.38} & \textbf{44.07 $\pm$ 0.75} & \textbf{47.5 $\pm$ 0.24} & \textbf{57.43 $\pm$ 0.31}\\
\midrule
Balanced Softmax & 78.03 $\pm$ 0.28&  81.63 $\pm$ 0.39& 88.10 $\pm$ 0.32& 42.11 $\pm$ 0.70& 46.79 $\pm$ 0.24& 58.06 $\pm$ 0.40\\
\textbf{+ Ours} & \textbf{79.05 $\pm$ 0.20} & \textbf{82.76 $\pm$ 0.52} & \textbf{88.89 $\pm$ 0.21} & \textbf{42.86 $\pm$ 0.27} & \textbf{47.28 $\pm$ 0.58} & \textbf{58.80 $\pm$ 0.72}\\
\midrule
SSP & 74.58 $\pm$ 0.16&  79.20 $\pm$ 0.43& 88.50 $\pm$ 0.24& 43.00 $\pm$ 0.51& 47.04 $\pm$ 0.60& 59.08 $\pm$ 0.46\\
\textbf{+ Ours} & \textbf{79.38 $\pm$ 0.65} & \textbf{82.18 $\pm$ 0.33} & \textbf{88.80  $\pm$ 0.43} & \textbf{43.57 $\pm$ 0.29} & \textbf{48.66 $\pm$ 0.57} & \textbf{59.78 $\pm$ 0.91}\\
\bottomrule
\end{tabular}
}
}
\vspace{-10pt}
\end{table*}

\subsection{Comparison methods}
In this section, we verify that Open-sampling can boost the standard training and several state-of-the-art techniques by integrating Open-sampling with the following methods: 
1) Standard: all the examples have the same weights; by default, we use standard cross-entropy loss. 
2) SMOTE \cite{chawla2002smote}: a variant of re-sampling with data augmentation. 
3) CB-RW \cite{cui2019class}: training examples are re-weighted according to the inverse of the effective number of samples in each class, defined as $(1-\beta^{n_i})/(1-\beta)$. 
4) CB-Focal \cite{cui2019class}: the CB method is combined with Focal loss.
5) M2m \cite{kim2020m2m}: an over-sampling method with adversarial examples.
6) LDAM-RW \cite{cao2019learning}: the method derives a generalization error bound for the imbalanced training and uses a margin-aware multi-class weighted cross entropy loss. 
7) LDAM-DRW \cite{cao2019learning}: the network is trained with LDAM loss and deferred re-balancing training. 
8) Balanced Softmax \cite{Ren2020balms}: the method derives a Balanced Softmax function from the probabilistic perspective that explicitly models the test-time label distribution shift. 
9) SSP \cite{yang2020rethinking}: the method uses self-supervised learning to pre-train the network on the auxiliary dataset before standard training. 
Here, we do not expect vanilla Open-sampling to achieve state-of-the-art results compared with many complicated methods, our method can be still a promising option in the family of class-imbalanced learning methods, because it can outperform existing data re-balancing methods and improve existing state-of-the-art methods.

 \begin{table}[!t]
\centering
\caption{Classification accuracy (\%) on CIFAR-10 under the setting of Balance Softmax \cite{Ren2020balms}. The bold indicates the improved results by integrating our regularization. Baseline results are taken from Balance Softmax\cite{Ren2020balms}. ``
$\star$'' indicates decoupled training. }
\label{tab:cifar10_new_results}
\renewcommand\arraystretch{0.9}
\resizebox{0.45\textwidth}{!}{
\setlength{\tabcolsep}{3mm}{
\begin{tabular}{c|ccc}
\toprule
Imbalance Factor & 200 & 100 & 10 \\
\midrule
Standard & 71.2 & 77.4 & 90.0  \\
CB-RW & 72.5 & 78.6 & 90.1 \\
LDAM-RW & 73.6 & 78.9 & 90.3  \\
Equalization Loss \cite{tan2020equalization} & 74.6 & 78.5 & 90.2  \\
Balanced Softmax & 79.0 & 83.1 & 90.9  \\
Ours & 80.6 & 83.6 & 90.6  \\
cRT$^{\star}$ \cite{kang2019decoupling} & 76.6 & 82.0 & 91.0  \\
LWS$^{\star}$ \cite{kang2019decoupling} & 78.1 & 83.7 & \textbf{91.1}  \\
Ours$^{\star}$ & \textbf{81.1} & \textbf{84.9} & 90.9  \\

\bottomrule
\end{tabular}
}
}
\vspace{-20pt}
\end{table}

\subsection{Main results}

\textbf{Results on long-tailed CIFAR.} Extensive experiments are conducted on long-tailed CIFAR datasets with three different imbalance ratios: 10, 50, and 100. We use 300K Random Images\footnote{The dataset is published on \url{https://github.com/hendrycks/outlier-exposure}.} \cite{hendrycks2019oe} as the open-set auxiliary dataset. The test accuracy of ResNet-32 \cite{he2016deep} on long-tailed CIFAR datasets are reported in Table \ref{tab:cifar10_results}. The results show that our method can achieve impressive improvements on the standard training method. Especially for long-tailed CIFAR-10 with imbalance ratio 100, an extreme imbalance case, the vanilla Open-sampling can significantly outperform the standard baseline by 8.39\%. Besides, incorporating our method into existing state-of-the-art methods can consistently improve their performance under various imbalance ratios.

We also conduct experiments on long-tailed CIFAR-10 following the training setting of Balanced Softmax \cite{Ren2020balms} and present the results in Table \ref{tab:cifar10_new_results}. Under this setting, our proposed method achieves the best performance among the end-to-end methods and could also improve the performance of decoupled training \cite{kang2019decoupling}.

\begin{figure}[!t]
    \centering
    \includegraphics[width=0.35\textwidth]{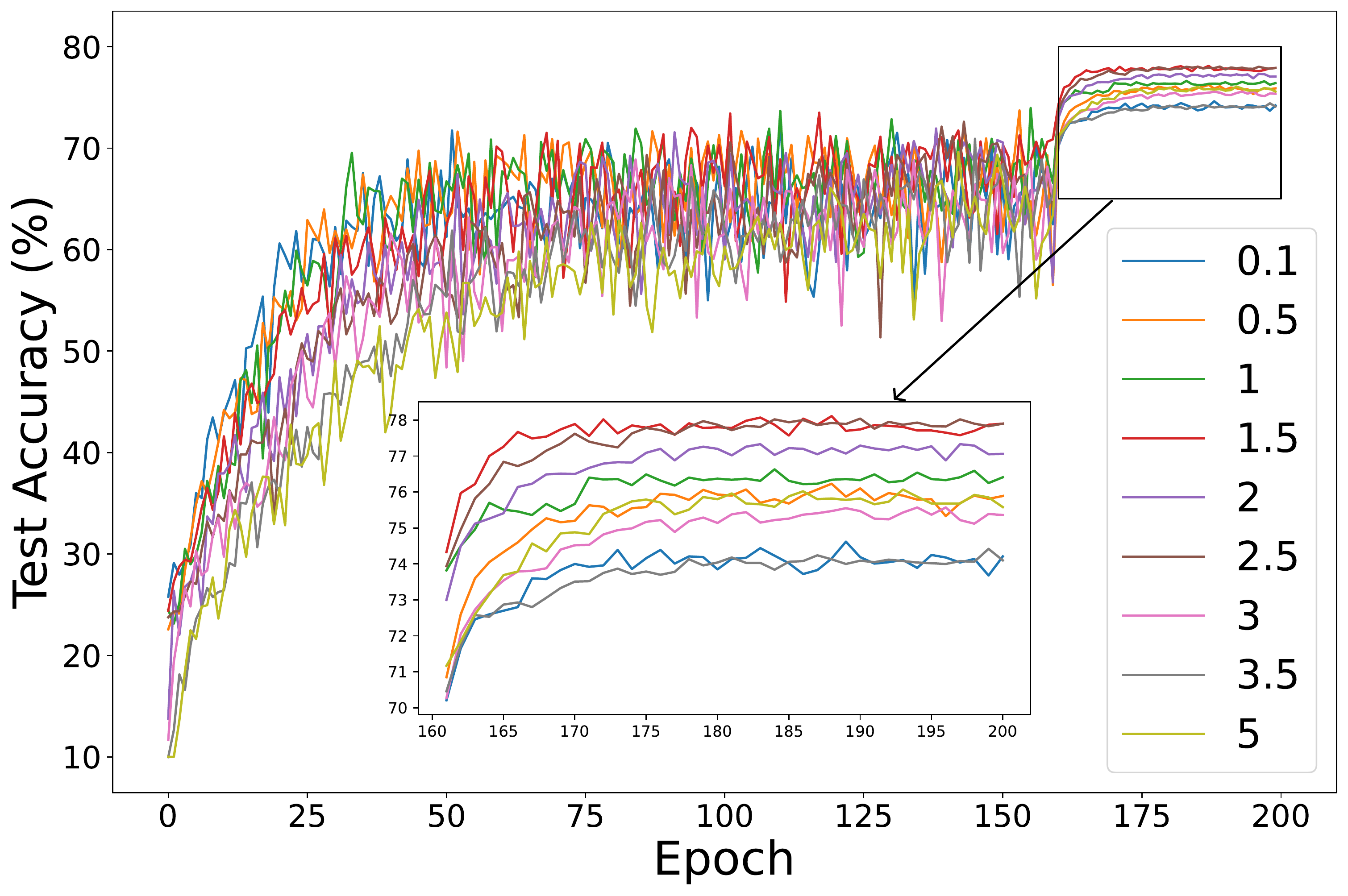}
    \caption{Results of sensitivity analysis on long-tailed CIFAR-10 with various values for $\eta$.}
    \label{fig:sensitivity}
    \vspace{-15pt}
\end{figure}

To further clarify the influence of $\eta$, we present a sensitivity analysis on long-tailed CIFAR-10 dataset with imbalance ratio 100 in Figure \ref{fig:sensitivity}. We highlight the differences in the trend of test accuracy after the decay of learning rate at the 160th epoch. From the figure, we can observe that with a proper value of $\eta$ like 1.5, the generalization performance can be largely improved by our proposed regularization. The result verifies that our regularization is an effective method to improve the generalization performance in long-tailed imbalanced learning.

\textbf{Results on CelebA-5.} We further verify the effectiveness of our method on real-world class-imbalanced datasets. The CelebA dataset has a long-tailed label distribution and the test set has a balanced label distribution. We use 300K Random Images \cite{hendrycks2019oe} as the open-set auxiliary dataset. Table \ref{tab:celeba_results} summarizes test accuracy on the CelebA-5 dataset. In particular, the proposed method outperforms existing data-rebalancing methods and is able to consistently improve the existing state-of-the-art methods in test accuracy.

\textbf{Results on Places-LT.} For Places-LT, we follow the protocol of Decoupled training \cite{kang2019decoupling} and start from a ResNet-152 \cite{he2016deep} backbone pre-trained on the ImageNet dataset \cite{ILSVRC15}. We fine-tune the backbone with Instance-balanced sampling for representation learning and then re-train the classifier with our proposed algorithm as decoupled training. Here, we use Places-extra69 \cite{zhou2017places} as the open-set auxiliary dataset. Table \ref{tab:places_results} summarizes test accuracy on the Places-LT dataset. The proposed method achieves the best overall performance, show that our method and decoupled training scheme are two orthogonal components and Open-sampling is applicable for large-scale datasets.

\textbf{OOD detection under class imbalance}. Out-of-distribution (OOD) detection is an essential problem for the deployment of deep learning especially in safety-critical applications \cite{hendrycks2019oe, liu2020energy, liang2017enhancing, yang2021oodsurvey} and it would be more difficult when the training dataset exhibits long-tailed distributions. Outlier Exposure (OE) \cite{hendrycks2019oe} is a commonly-used method to improve anomaly detection performance with auxiliary dataset. Specifically, the training objective of OE under class imbalance is $\mathbb{E}_{(x, y) \sim \mathcal{D}_{\mathrm {train}}}\left[-\log f_{y}(x)\right]+\lambda \mathbb{E}_{x \sim \mathcal{D}_{\mathrm{out}}}[H(P(Y) ; f(x))]$, where $H$ is the cross entropy and $P(Y)$ is the label prior of the original training dataset. Following OE \cite{hendrycks2019oe}, we consider the maximum softmax probability (MSP) baseline \cite{hendrycks2016baseline}. Here, we conduct experiments on long-tailed CIFAR10 with imbalance rate 100 to verify the advantage of Open-sampling on OOD detection under class imbalance. Table \ref{tab:ood} presents the test accuracy and the average performance over the three types of noises and seven OOD test datasets. We can observe that our method achieves impressive improvement on both the test accuracy and the detection performance.

\begin{table}[!t]
\centering
\caption{Classification accuracy (\%) on CelebA-5 with ResNet-32. ``$\dagger$'' indicates the reported results are obtained from \cite{kim2020m2m}. The shadow indicates the improved results.}
\label{tab:celeba_results}
\renewcommand\arraystretch{1.8}
\resizebox{0.45\textwidth}{!}{
\setlength{\tabcolsep}{1mm}{
\begin{tabular}{cc|cc|cc}
\toprule
Method & Accuracy & Method & Accuracy & Method & Accuracy\\
\midrule
Standard & 72.7 & M2m $^{\dagger}$& 75.6 & LDAM-DRW & 74.5  \\
SMOTE $^{\dagger}$& 72.8 & \CC\ \textbf{Ours} & \CC\ \textbf{76.8} &  \CC\ \textbf{ LDAM-DRW + Ours} & \CC\ \textbf{76.9}\\
CB-RW& 73.6 &  LDAM-RW & 73.1 & Balanced Softmax & 76.4\\
CB-Focal & 74.2 &  \CC\ \textbf{LDAM + Ours} & \CC\ \textbf{75.8} &\CC\ \textbf{Balanced Softmax + Ours} & \CC\ \textbf{78.6}  \\
\bottomrule
\end{tabular}
}
}
\vspace{-5pt}
\end{table}

\begin{table}[!t]
\centering
\caption{Top-1 accuracy (\%) on Places-LT with
an ImageNet pre-trained ResNet-152. Baseline results are taken from original papers. ``DT'' indicates decoupled training. }
\label{tab:places_results}
\renewcommand\arraystretch{1.2}
\resizebox{0.45\textwidth}{!}{
\setlength{\tabcolsep}{2mm}{
\begin{tabular}{c|ccc|c}
\toprule
Method & Many & Medium & Few & Overall \\
\midrule
Lifted Loss \cite{oh2016deep} & 41.1 & 35.4 & 24.0 & 35.2  \\
Focal Loss & 41.1 & 34.8 & 22.4 & 34.6 \\
Range Loss \cite{zhang2017range} & 41.1 & 35.4 & 23.2 & 35.1  \\
OLTR \cite{openlongtailrecognition} & 44.7 & 37.0 & 25.3 & 35.9  \\
cRT$^{\star}$ \cite{kang2019decoupling} & 42.1 & 37.6 & 24.9 & 36.7  \\
LWS$^{\star}$ \cite{kang2019decoupling} & 40.6 & 39.1 & 28.6 & 37.6  \\
Ours$^{\star}$ & 42.9 & \textbf{39.3} & 26.8 & \textbf{38.2}  \\

\bottomrule
\end{tabular}
}
}
\vspace{-10pt}
\end{table}

\begin{table}[!t]
\centering
\renewcommand\arraystretch{1}
\caption{ OOD detection performance comparison on long-tailed CIFAR-10. All values are percentages and are averaged over the ten test datasets described in Appendix \ref{app:settings}. ``$\uparrow$'' indicates larger values are better, and ``$\downarrow$'' indicates smaller values are better. Bold numbers are superior results. Detailed results for each OOD test dataset can be found in Appendix \ref{app:results}.} 
\label{tab:ood}
\renewcommand\arraystretch{1.0}
\resizebox{0.45\textwidth}{!}{
\setlength{\tabcolsep}{2mm}{
\begin{tabular}{ccccc}
\toprule
 Method & Test Accuracy $\uparrow$ & FPR95 $\downarrow$	& AUROC $\uparrow$		& AUPR $\uparrow$	\\
\midrule
 MSP & 71.83 &  56.1 &  75.2 &  32.71\\
 OE & 66.74&  32.38 &  84.15 &  36.86 \\
Ours & \textbf{77.62}&  \textbf{20.68} &  92.40 &  58.38 \\
Ours ($\alpha=5$) & 75.16&  22.13 &  \textbf{94.26} &  \textbf{75.91} \\
\bottomrule

\end{tabular}
}
}
\vspace{-10pt}
\end{table}

\begin{figure*}
    \centering
    \begin{subfigure}[b]{0.32\textwidth}
        \centering
        \includegraphics[width=\textwidth]{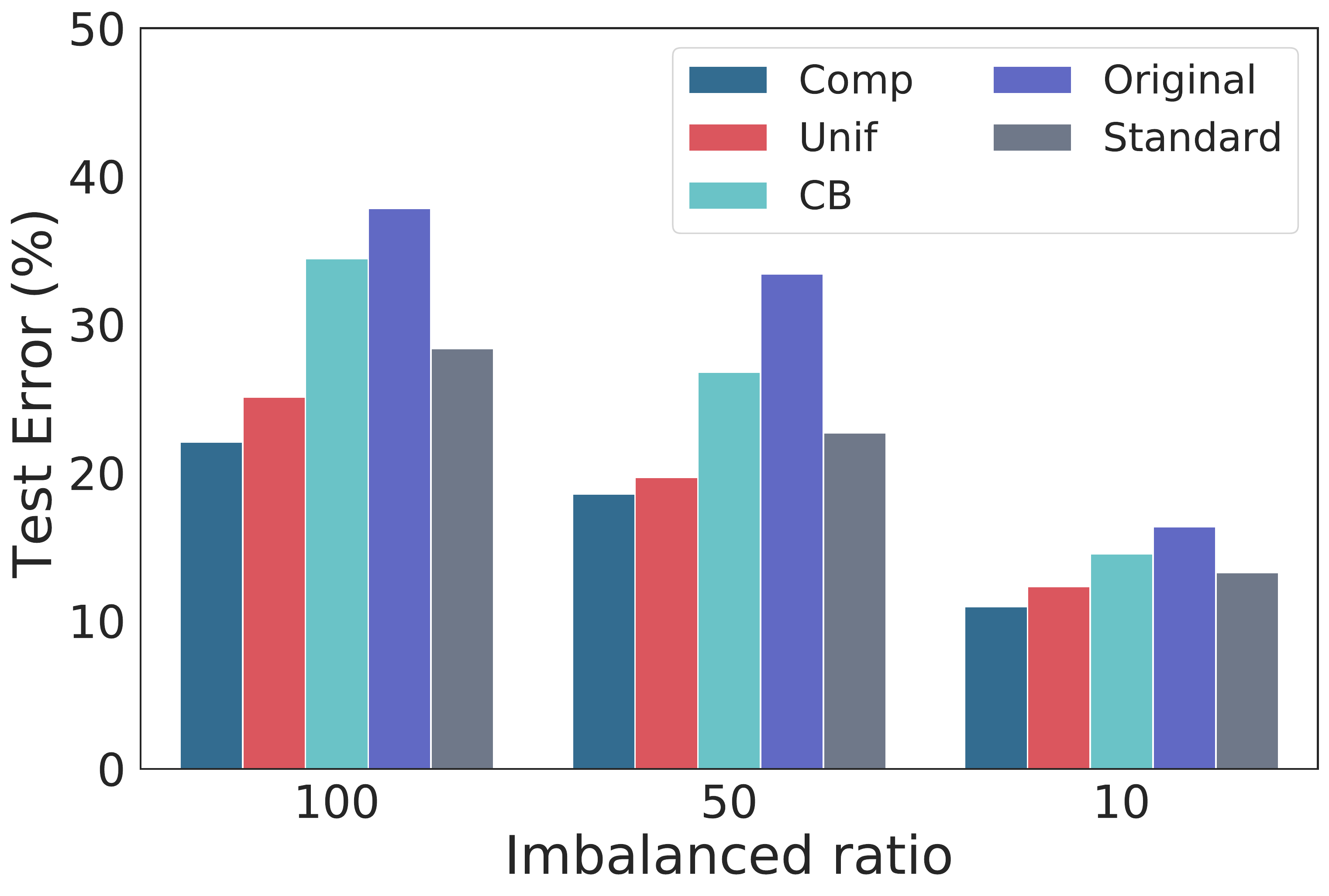}
        \caption{Label distribution.}
        \label{fig:distribution}
    \end{subfigure}
    \hfill
    \begin{subfigure}[b]{0.32\textwidth}
        \centering
        \includegraphics[width=\textwidth]{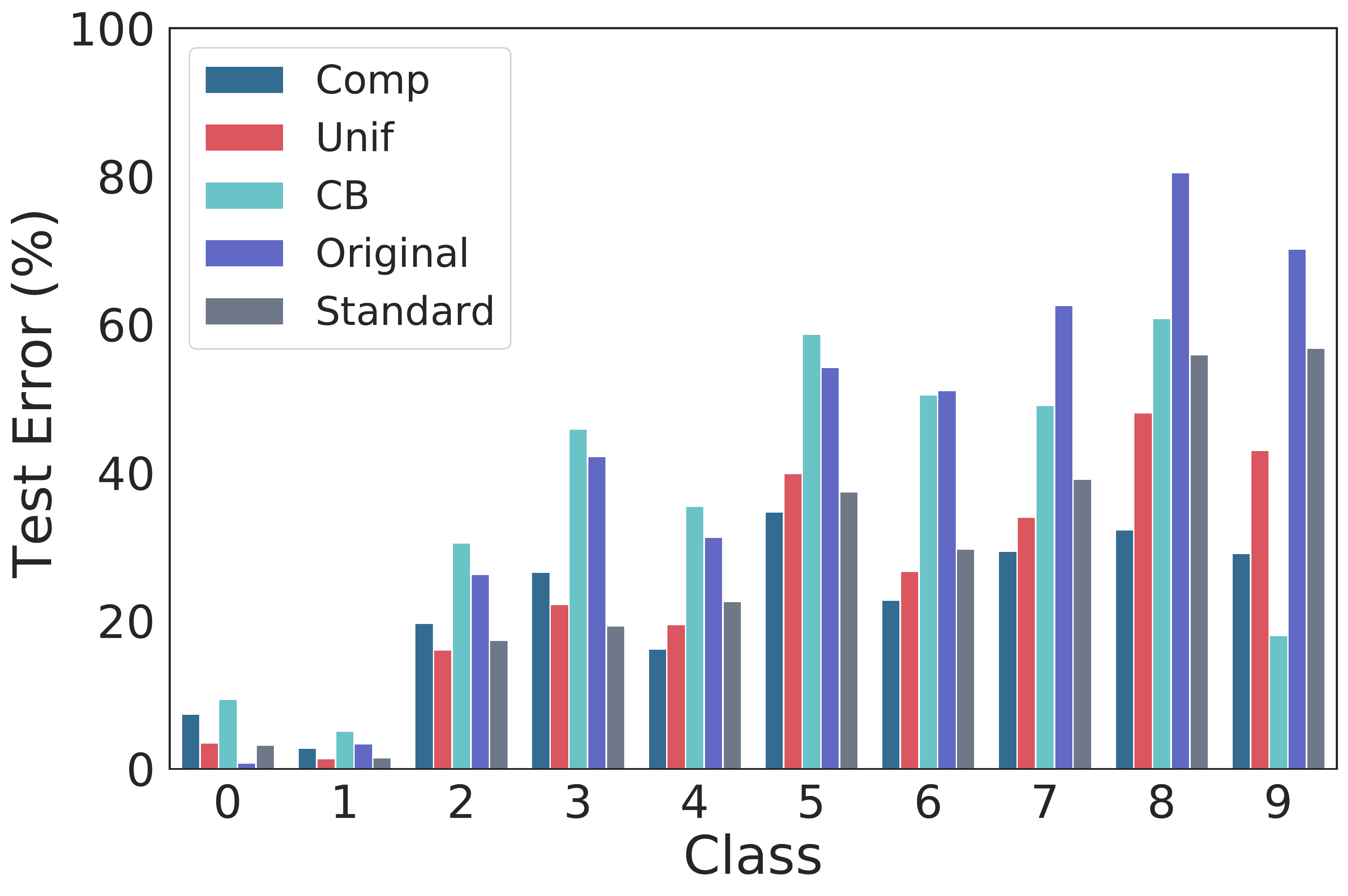}
        \caption{Label distribution.}
        \label{fig:test_error_class}
    \end{subfigure}
    \hfill
    \begin{subfigure}[b]{0.32\textwidth}
        \centering
        \includegraphics[width=\textwidth]{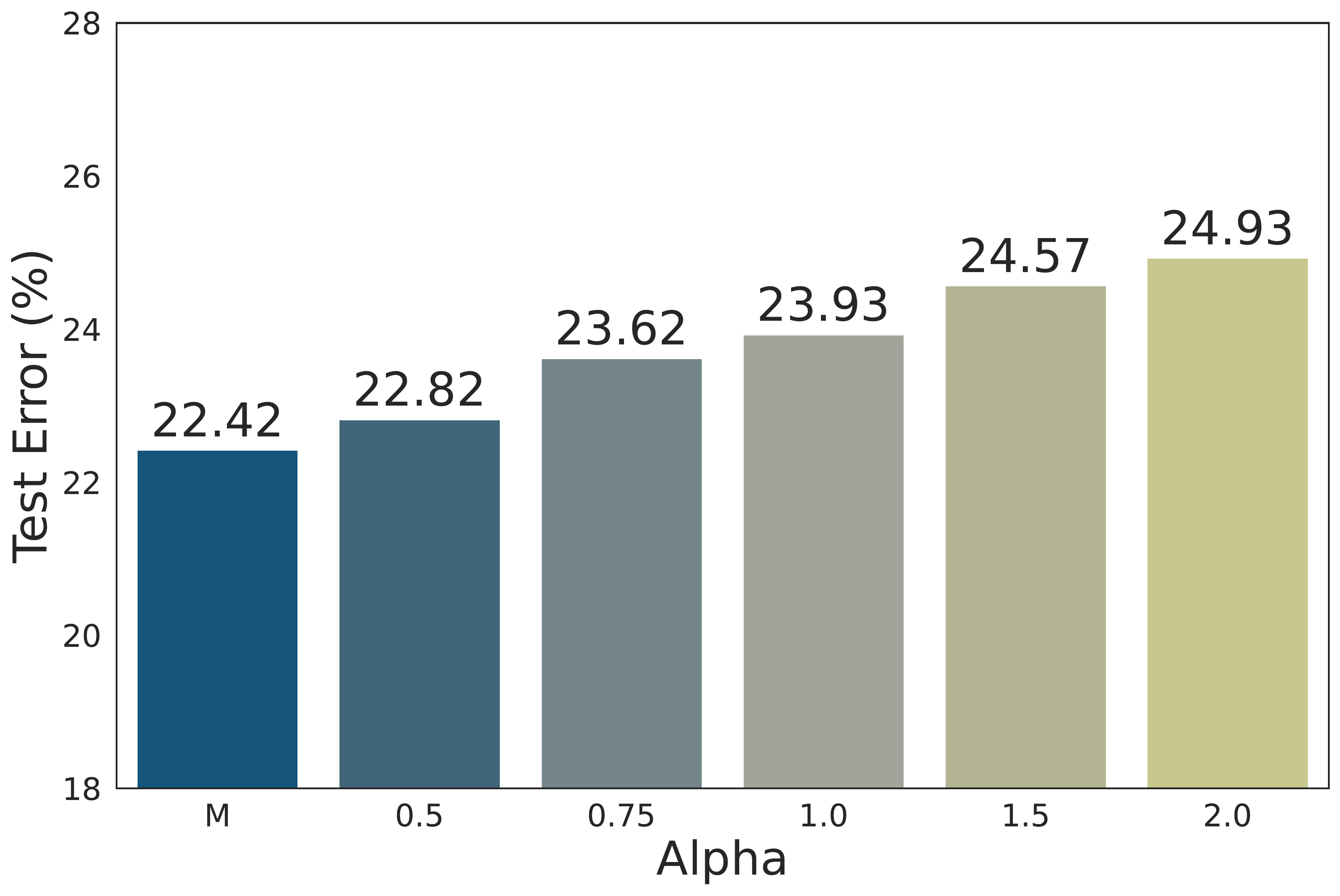}
        \caption{Alpha.}
        \label{fig:alpha}
    \end{subfigure}
    \hfill
    \begin{subfigure}[b]{0.32\textwidth}
        \centering
        \includegraphics[width=\textwidth]{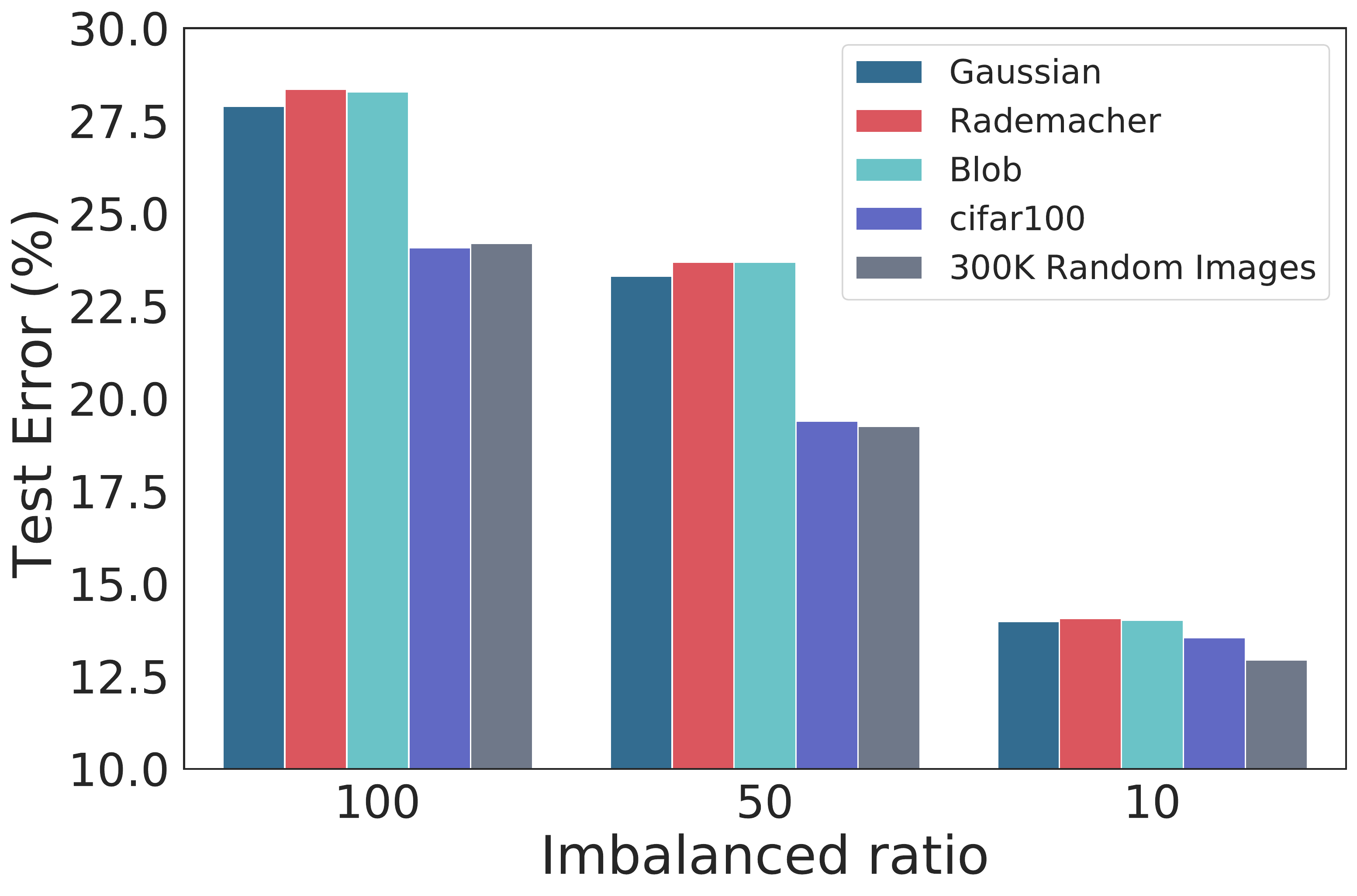}
        \caption{Auxiliary dataset.}
        \label{fig:aux_set}
    \end{subfigure}
    \hfill
    \begin{subfigure}[b]{0.32\textwidth}
        \centering
        \includegraphics[width=\textwidth]{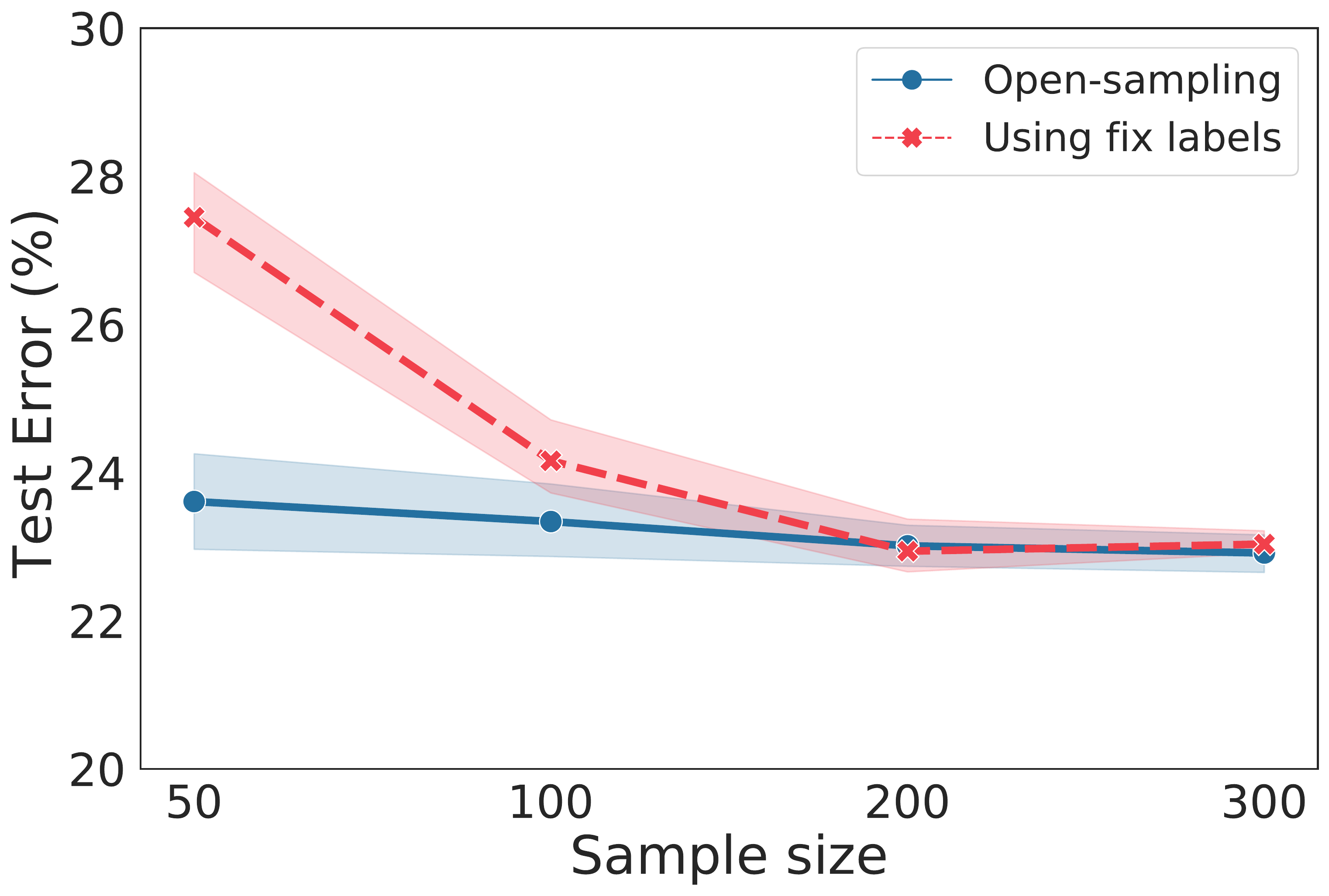}
        \caption{Sample size.}
        \label{fig:size}
    \end{subfigure}
    \hfill
    \begin{subfigure}[b]{0.32\textwidth}
        \centering
        \includegraphics[width=\textwidth]{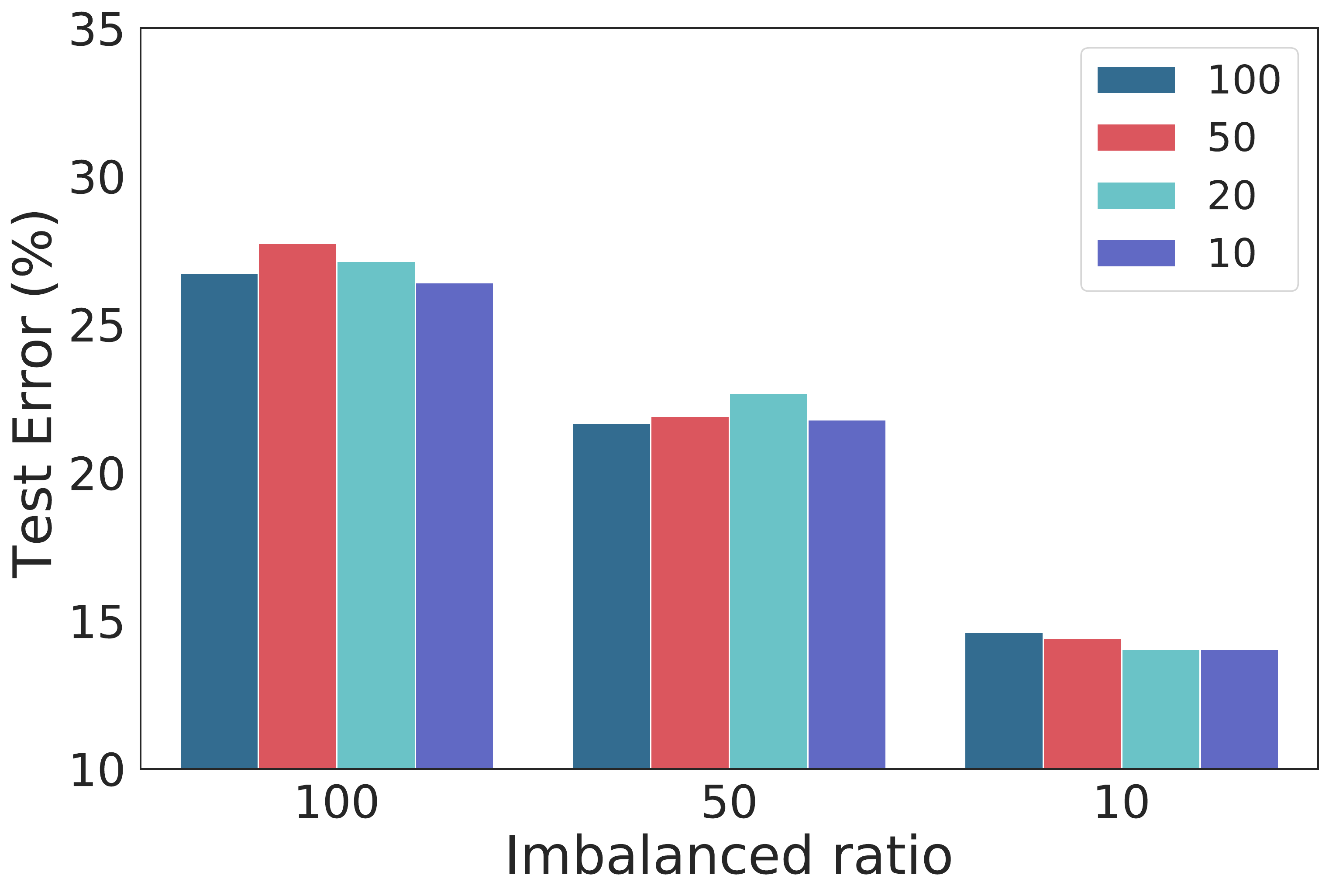}
        \caption{Number of Classes.}
        \label{fig:sub_class}
    \end{subfigure}
     \caption{Analytical experiments of Open-sampling on long-tailed CIFAR-10: (\ref{fig:distribution})(\ref{fig:test_error_class}) with various label distributions, (\ref{fig:alpha}) with various values of the $\alpha$, (\ref{fig:aux_set}) with various open-set auxiliary datasets, including simulated noise datasets and real-world datasets. All the datasets contain 50,000 instances. (\ref{fig:size}) with various sample sizes (K) of the auxiliary dataset, (\ref{fig:sub_class}) with various number of classes in the auxiliary datasets that are randomly sampled from CIFAR-100. Experiments in (\ref{fig:test_error_class}), (\ref{fig:alpha}), and (\ref{fig:size}) are conducted under the imbalance ratio 100. The y-axis represents the test error in all the figures. ``Standard" denotes the baseline with standard cross-entropy loss.}
     \label{fig:analysis}
    \vspace{-8pt}
\end{figure*}

\section{A Closer Look at Open-sampling under Class Imbalance}
\label{sec:analyze}
To provide a comprehensive understanding of the proposed method, we conduct a set of analyses in this section. Firstly, we compare our defined distribution with several alternative distributions to show the advantage of the Complementary Distribution in our method. Secondly, the effect of $\alpha$ in our method is thoroughly analyzed by empirical studies. Then, we present a guideline about how to choose or collect a suitable open-set auxiliary dataset for long-tailed imbalanced learning. Finally, we analyze the effect of the proposed method through the lens of decision boundaries.

\textbf{The advantage of Complementary Distribution.} In the proposed method, the labels of OOD instances from the auxiliary dataset are sampled from a random label distribution. For the random label distribution, we defined a Complementary Distribution in Proposition \ref{prop:complementary} and also presented several commonly used distributions, including uniform distribution (Unif), class balanced distribution (CB) \cite{cui2019class}, and the original class priors of the training dataset (Original). Here, we conduct experiments to compare the performance of the Open-sampling variants with different label distributions. The results in Figure \ref{fig:distribution} show that using the Complementary Distribution consistently achieves the best performance on the test set. In particular, using the uniform distribution can also improve the generalization performance while both CB and the original class priors deteriorate the performance of the neural networks.

To further understand why CB is not a good choice in our method, we present the per-class top-1 error on long-tailed CIFAR-10 in Figure \ref{fig:test_error_class}. Although using the CB distribution can achieve better performance on the smallest class, it downgrades the generalization performance on the other classes. The reason is that the CB distribution is far away from the uniform distribution, thereby introducing too much noise to the Bayes classifier. Different from CB, the Complementary Distribution is closer to a uniform distribution, making it achievable to re-balance the class priors while almost keeping non-toxicity of the open-set noisy labels.

Here, we also show the effect of $\alpha$ in Figure \ref{fig:alpha}. As analyzed in Section \ref{sec:method}, the larger the value of $\alpha$ is, the Complementary Distribution tends to be closer to a uniform distribution. Here, ``M'' denotes the default value of $\alpha = (\max_j \beta_j + \min_j \beta_j)$. Additionally, the MCD variant achieves $22.51\%$ on the test error. From Figure \ref{fig:alpha}, the test error presented a slightly upward trend with the increasing of the value of $\alpha$. The results verified that it is necessary to adopt a complementary distribution, instead of simply using a uniform distribution.

\begin{figure*}[!t]
    \centering
    \begin{subfigure}[b]{0.26\textwidth}
        \centering
        \includegraphics[height=4.0cm,width=5cm]{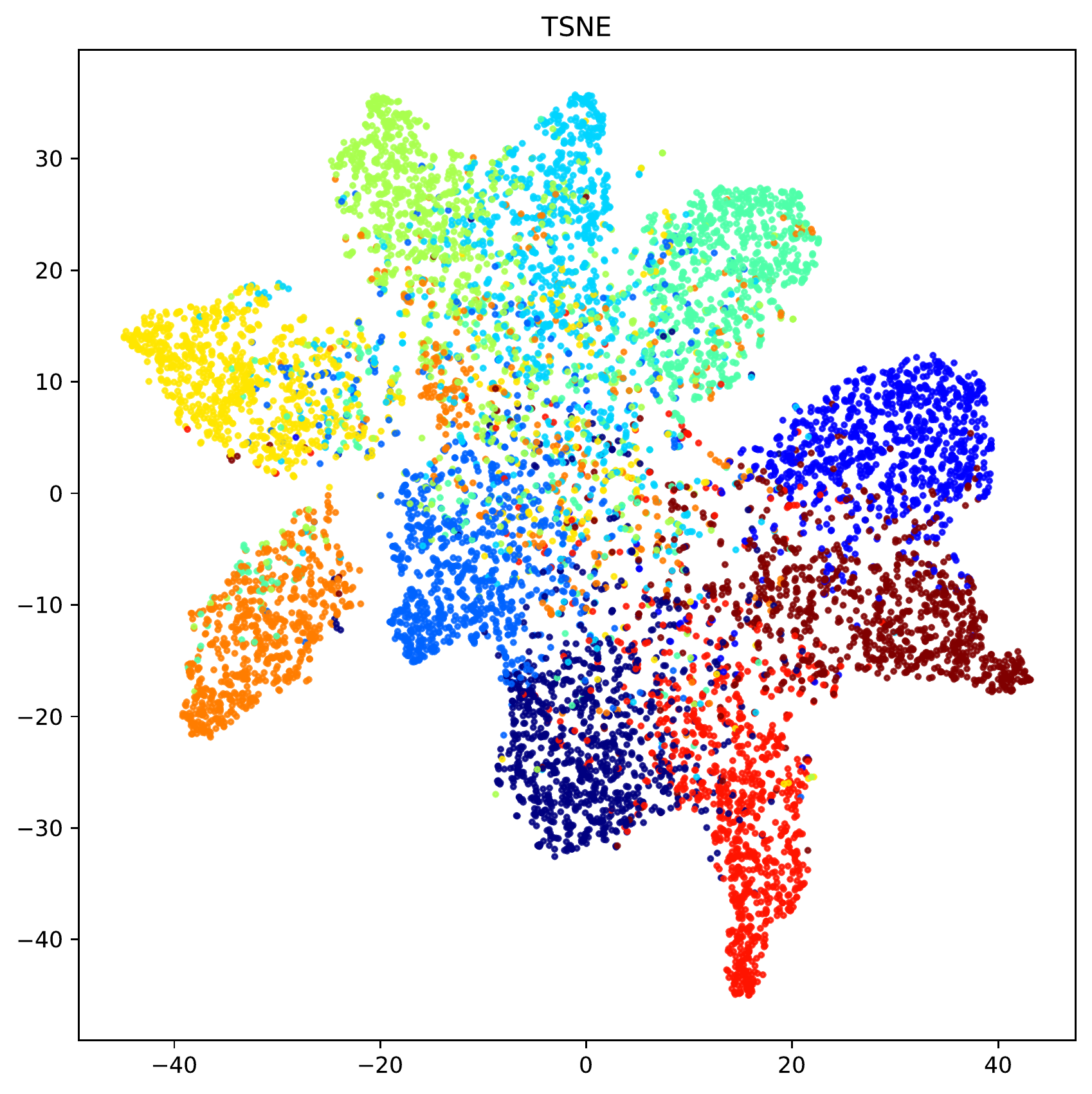}
        \caption{Standard}
        \label{fig:standard_tsne}
    \end{subfigure}
\quad\quad\quad
    \begin{subfigure}[b]{0.26\textwidth}
        \centering
        \includegraphics[height=4.0cm,width=5cm]{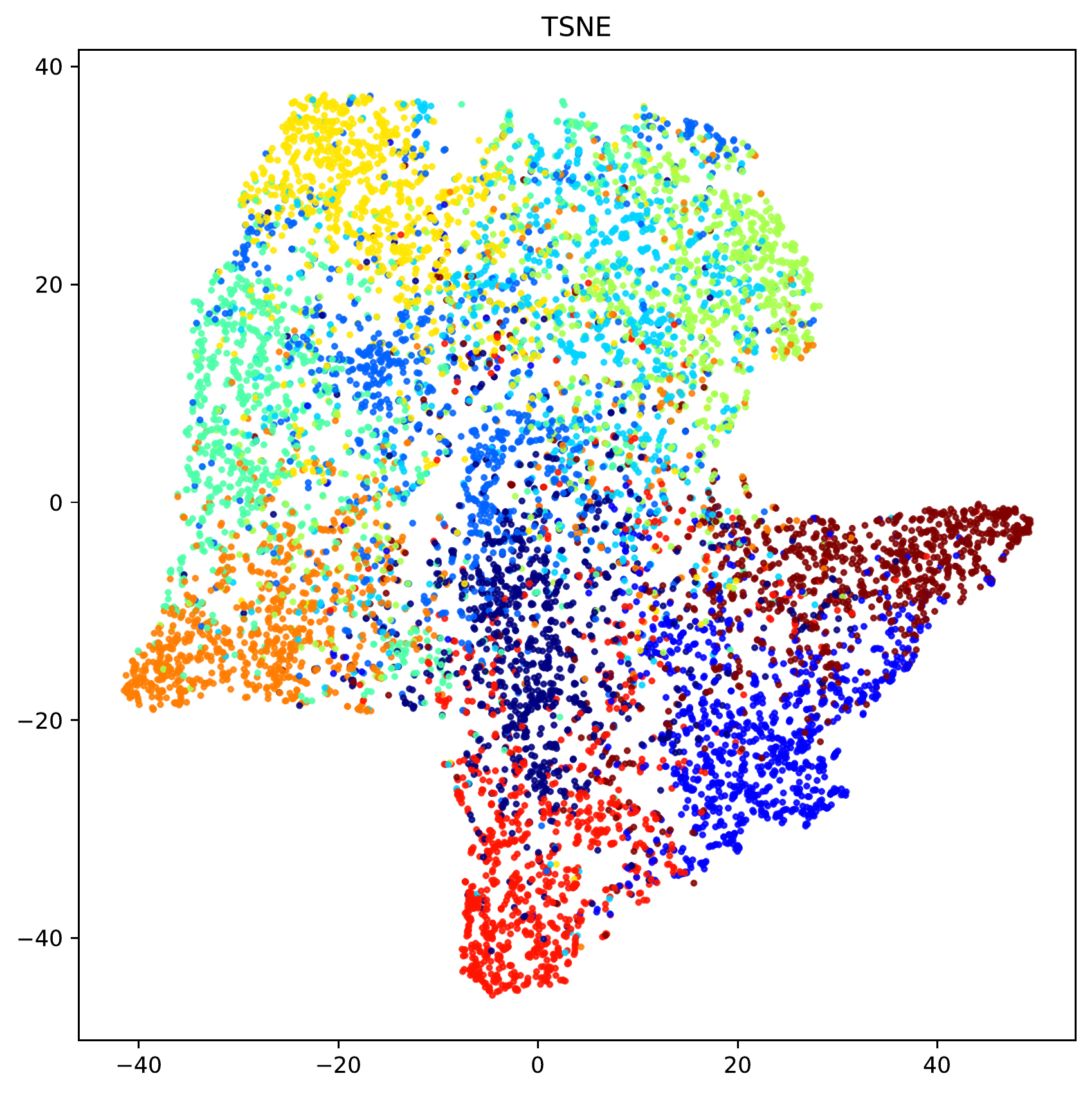}
        \caption{CB-Focal}
        \label{fig:cb_focal_tsne}
    \end{subfigure}
\quad\quad\quad
    \begin{subfigure}[b]{0.26\textwidth}
        \centering
        \includegraphics[height=4.0cm,width=5cm]{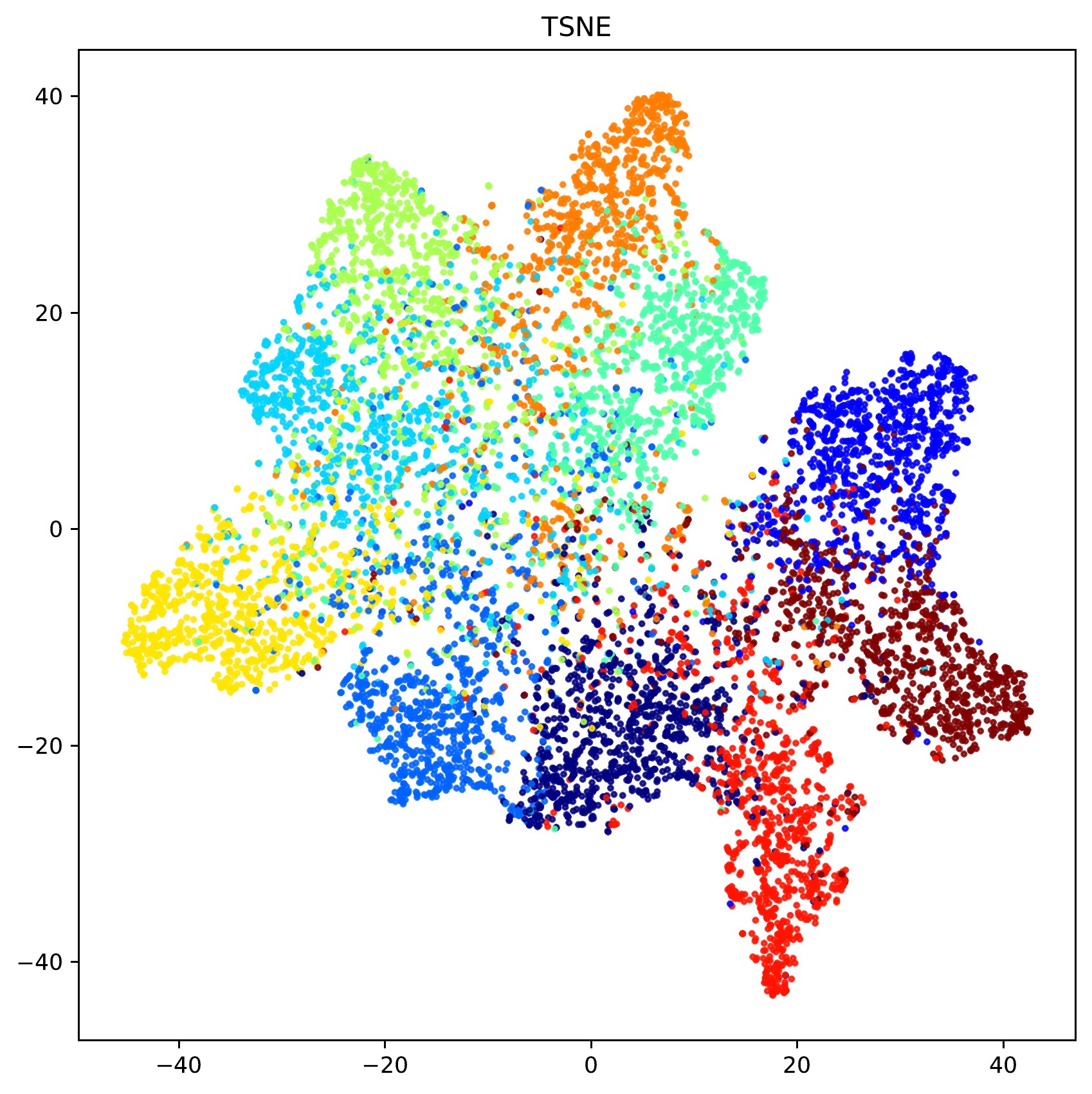}
        \caption{CB-Resampling}
        \label{fig:resample_tsne}
    \end{subfigure}
\quad\quad\quad
    \begin{subfigure}[b]{0.26\textwidth}
        \centering
        \includegraphics[height=4.0cm,width=5cm]{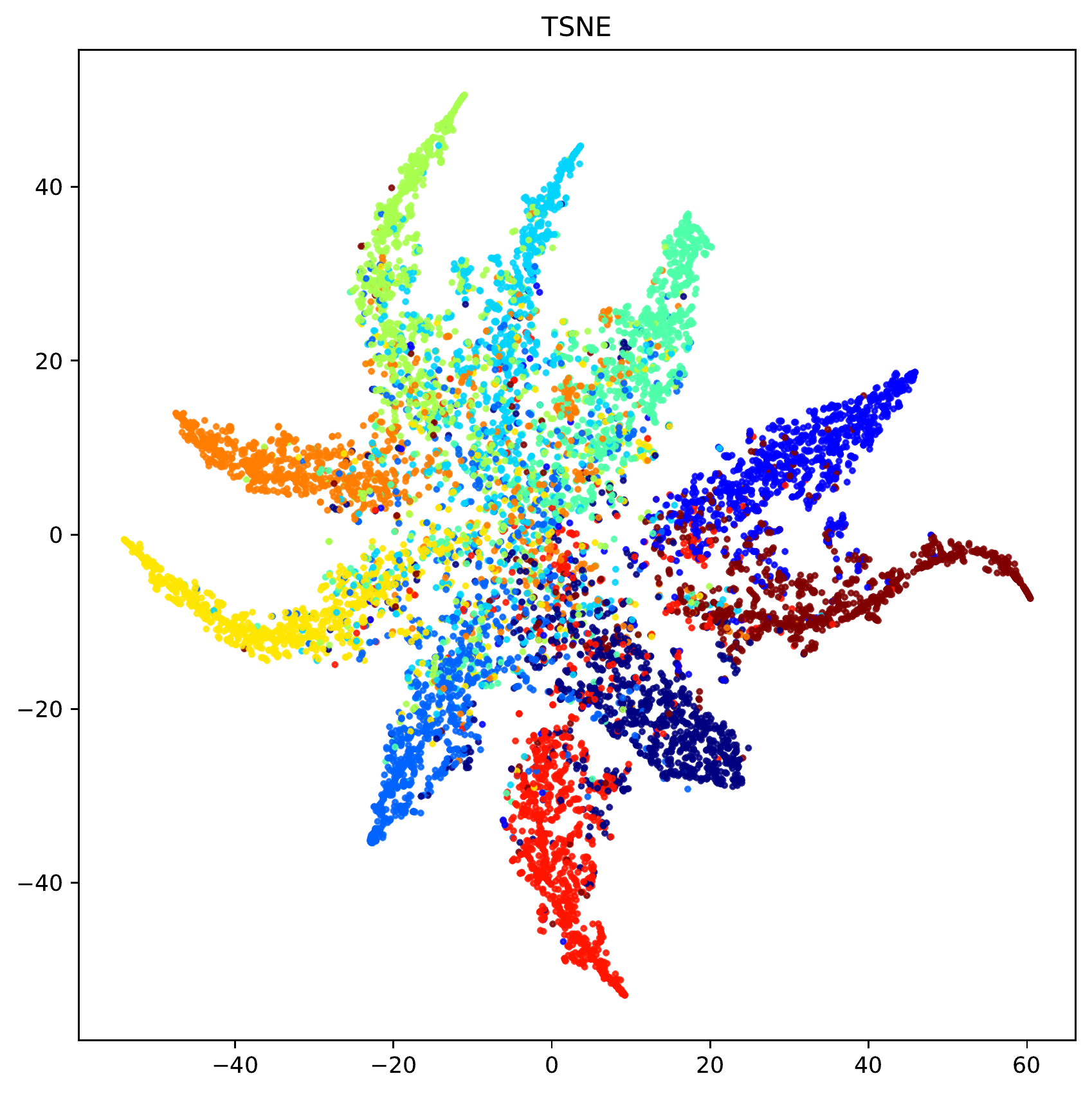}
        \caption{LDAM-RW}
        \label{fig:ldam_tsne}
    \end{subfigure}
\quad\quad\quad
    \begin{subfigure}[b]{0.26\textwidth}
        \centering
        \includegraphics[height=4.0cm,width=5cm]{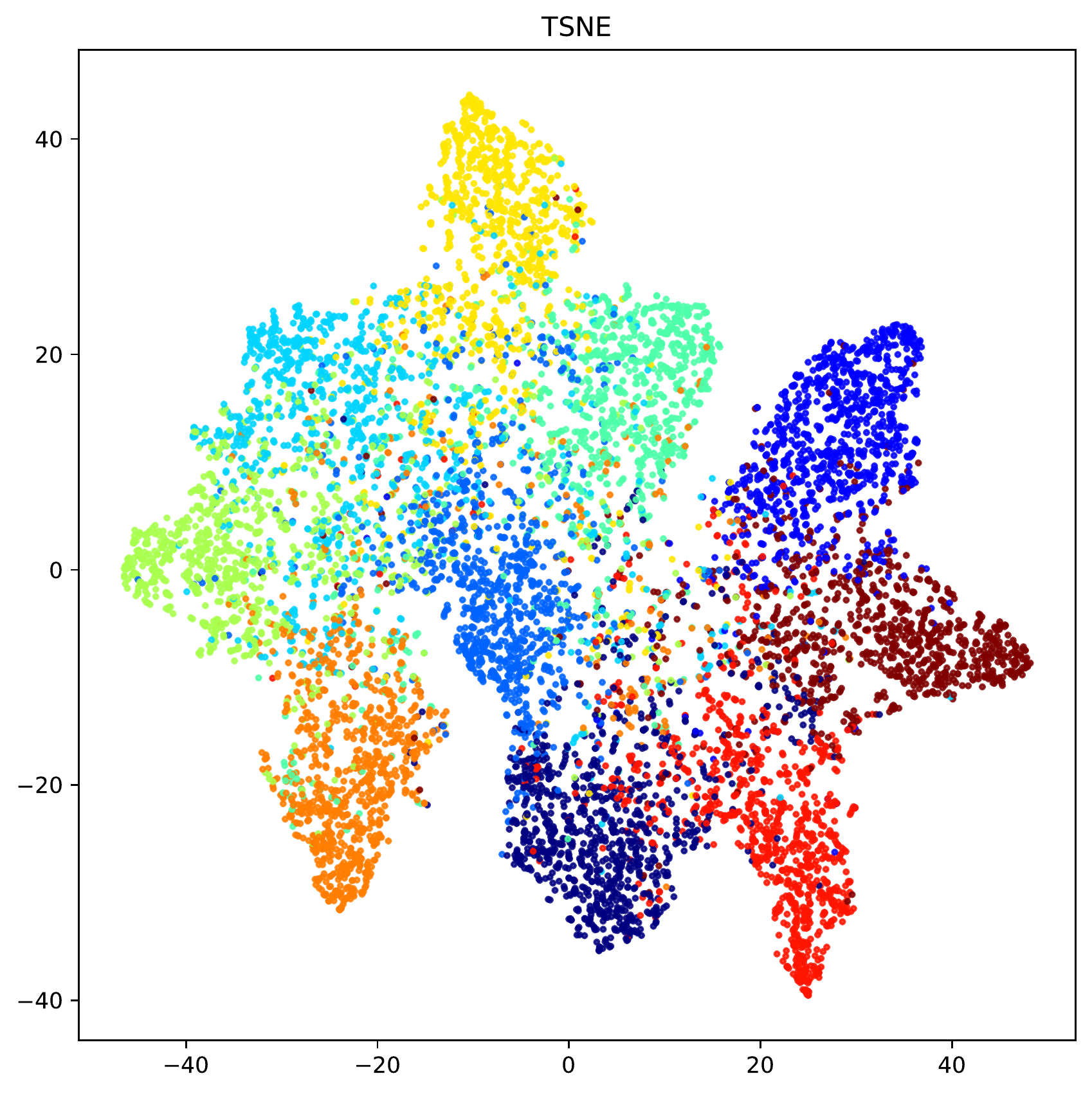}
        \caption{Balanced Softmax}
        \label{fig:bs_tsne}
    \end{subfigure}
\quad\quad\quad
    \begin{subfigure}[b]{0.26\textwidth}
        \centering
        \includegraphics[height=4.0cm,width=5cm]{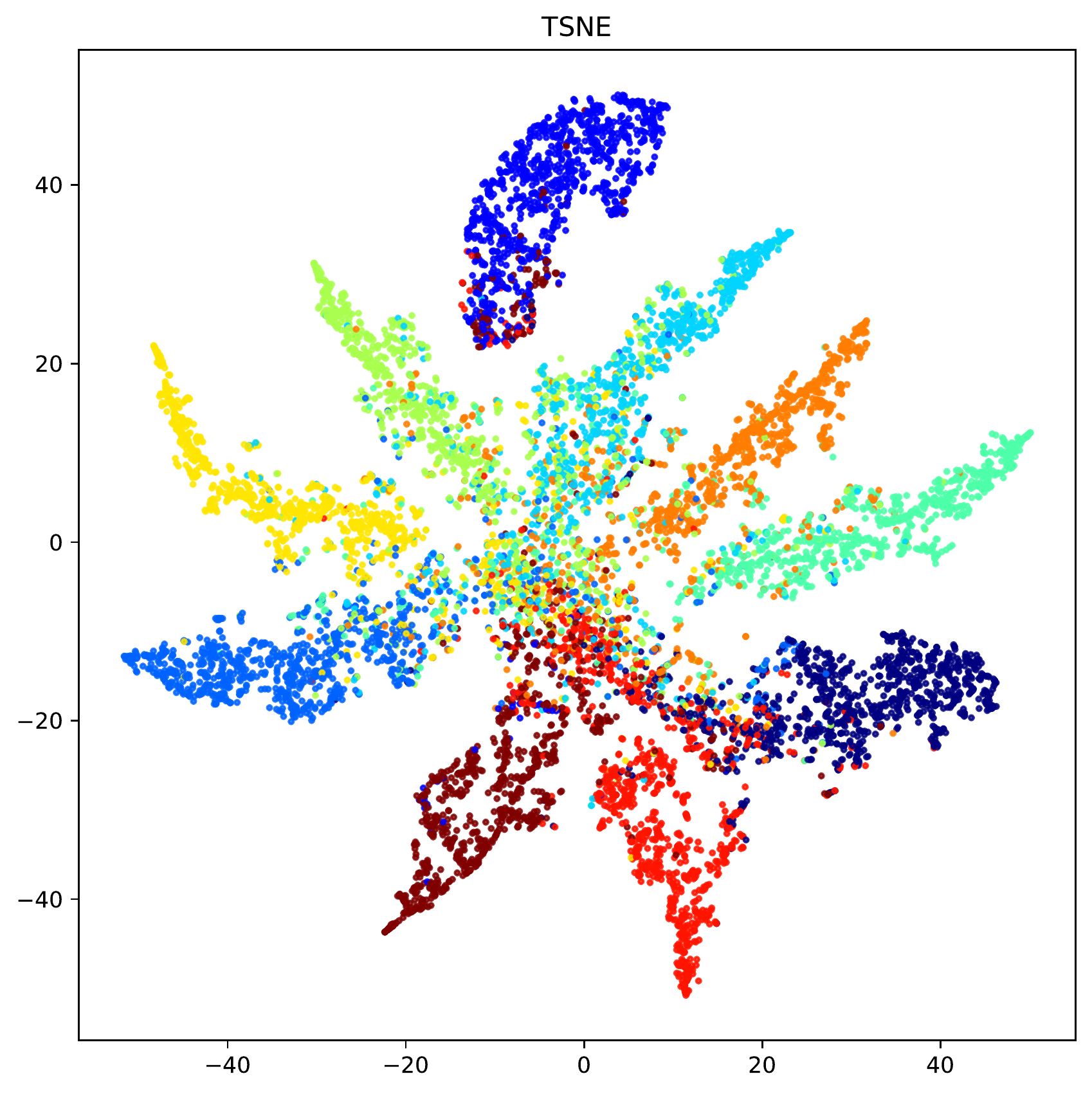}
        \caption{Ours}
        \label{fig:ours_tsne}
    \end{subfigure}
    \caption{t-SNE visualization of test set on long-tailed CIFAR-10 with imbalance ratio 100. We can observe that LDAM and our method appear to learn more separable representations than Standard training and the other algorithms.}
    \label{fig:tsne}
    \vspace{-10pt}
\end{figure*}

\textbf{The choices of open-set auxiliary dataset.} With proper label distribution, can any OOD dataset be used to improve Long-tail imbalanced learning? In Figure \ref{fig:aux_set}, we show the test error on long-tailed CIFAR-10 with imbalance ratio 100 using Open-sampling with different auxiliary datasets. We can observe that using simple noises like Gaussian noise could not achieve similar performance than those of real-world datasets, \emph{e.g.}, CIFAR100 and 300K Random Images \cite{hendrycks2019oe}. 

The sample size is another important factor for the open-set auxiliary dataset. In Figure \ref{fig:size}, we show that the performance of Open-sampling would be slightly better with a larger sample size of the auxiliary dataset. It is worth noting that the phenomenon is not consistent with that of \cite{wei2021open}, where using larger open-set auxiliary dataset could not improve the performance on learning with noisy labels. Here, we also conduct experiments using a variant of Open-sampling with fixed labels for OOD instances. We can observe that the variant performs poorly with a small auxiliary dataset, but the performance can be significantly improved by increasing the sample size of auxiliary dataset. In particular, the variant could achieve nearly the same improvement as the proposed Open-sampling method, with a large enough sample size.


In addition, we also find that the diversity of the open-set auxiliary dataset is unimportant for the effectiveness of Open-sampling. We conduct experiments on long-tailed CIFAR-10 with imbalance ratio 100 and use the subset of CIFAR100 with different number of classes as the open-set auxiliary dataset. The sample size of these subsets are fixed as 5,000. The results of using different subsets are almost the same, achieving 73.28\% test accuracy. The results are consistent with that in Figure \ref{fig:aux_set}, where using CIFAR100 as open-set auxiliary dataset can achieve almost the same improvements as 300K Random Images \cite{hendrycks2019oe}.

\textbf{The effect of Open-sampling as regularization.} In addition to re-balancing the class prior, what is the effect of Open-sampling as a regularization method? 
To gain additional insight, we look at the t-SNE projection of the learnt representations for different algorithms in Figure \ref{fig:tsne}. For each method, the projection is performed over test data. The figures show that the decision boundaries of Open-sampling and LDAM \cite{cao2019learning} are much clearer than those of the other methods. The phenomenon illustrates that Open-sampling can encourage the minority classes to have a larger margin, which is similar to the effect of the LDAM loss \cite{cao2019learning}. As shown in Section \ref{sec:exp}, our method could still boost the performance of the LDAM method, which demonstrates the differences between our works.

\section{Related Work}

\textbf{Re-sampling.} Re-sampling methods aims to re-balance the class priors of the training dataset. Under-sampling methods remove examples from the majority classes, which is infeasible under extremely data imbalanced settings \cite{he2009learning, japkowicz2002class}. 
The over-sampling method adds repeated samples for the minority classes, usually causing over-fitting to the minority classes \cite{buda2018systematic, byrd2019effect, shen2016relay}. 
Some methods utilize synthesized in-distribution samples to alleviate the over-fitting issue but introduce extra noise \cite{chawla2002smote,he2008adasyn,kim2020m2m}. 
In contrast to in-distribution samples used in existing methods, our approach exploits OOD instances to re-balance the class priors of the training dataset.

\textbf{Re-weighting.} Re-weighting methods propose to assign adaptive weights for different classes or samples. Generally, the vanilla scheme re-weights classes proportionally to the inverse of their frequency \cite{huang2016learning}. Focal loss \cite{lin2017focal} assigns low weights to the well-classified examples. Class-balanced loss \cite{cui2019class} proposes to re-weight by the inverse effective number of samples. However, these re-weighting methods tend to make the optimization of DNNs difficult under extremely data imbalanced settings \cite{wang2017learning, mikolov2013distributed}.

\textbf{Other methods for long-tailed datasets}.  In addition to the data re-balancing approaches, some other solutions are also applied for class-imbalanced learning, including transfer-learning based methods \cite{yin2019feature, jamal2020rethinking}, two-stage training methods\cite{kang2019decoupling, zhong2021mislas, zhang2021distribution, openlongtailrecognition}, training objective based methods \cite{cao2019learning, Ren2020balms, hong2021disentangling}, expert methods \cite{wang2021longtailed} and self-supervised learning methods \cite{yang2020rethinking, cui2021parametric}. For example, transfer-learning based methods addressed the class-imbalanced issue by transferring features learned from head classes with abundant training instances to under-represented tail classes \cite{yin2019feature, jamal2020rethinking}. Two-stage training methods proposed to apply decoupled training where the classifier is re-balanced during the fine-tuning stage \cite{kang2019decoupling, zhong2021mislas, zhang2021distribution, openlongtailrecognition}. Generally, our proposed method is complementary to these existing methods and our method can further improve their performance, which are explicitly shown in our experiments.

\textbf{Utilizing auxiliary dataset}. In the deep learning community, auxiliary dataset is utilized in various contexts, e.g., adversarial machine learning \cite{madry2018towards,lee2021removing} and weakly supervised learning \cite{wei2020combating, wei2020metainfonet, zhu2020second, cheng2021learningsieve, wang2021policy, zhu2021clusterability, wei2022learning, zhu2022detect, zhu2022beyond}. For example, OE \cite{hendrycks2019oe} regularized the network to give conservative predictions on OOD instances. OAT \cite{lee2021removing} utilized OOD instances to improve generalization in adversarial training. ODNL \cite{wei2021open} proposed to use open-set auxiliary dataset to prevent the model from over-fitting inherent noisy labels. In long-tailed imbalanced learning, recent work \cite{yang2020rethinking} introduced unlabeled-in-distribution data to compensate for the lack of training samples. Another work \cite{su2021realistic} adopted semi-supervised learning techniques to improve performance on long-tailed datasets with out-of-class images from related classes, which are different from out-of-distribution data in our method. To the best of our knowledge, we are the first to explore the benefits of \emph{out-of-distribution data} in learning with long-tailed datasets. 

\textbf{OOD detection.} OOD detection is an essential building block for safely deploying machine learning models in the open world \cite{hendrycks2016baseline, hendrycks2019oe, liu2020energy, wei@logitnorm}. A common baseline for OOD detection is the softmax confidence score \cite{hendrycks2016baseline}. It has been theoretically shown that neural networks with ReLU activation can produce arbitrarily high softmax confidence for OOD inputs \cite{hein2019relu}. To improve the performance, previous methods have explored using artificially synthesized data from GANs \cite{goodfellow2014generative} or unlabeled data \cite{hendrycks2019oe, lee2017training} as auxiliary OOD training data. Energy scores are shown to be better for distinguishing in- and out-of-distribution samples than softmax scores \cite{liu2020energy}. As a side effect, our method could also achieve superior performance in OOD detection tasks even if the labels of training dataset are noisy.

\section{Conclusion}
In this paper, we propose a simple yet effective method termed Open-sampling, by introducing OOD instances to re-balance the class priors of the training dataset. To the best of our knowledge, our method is the first to utilize OOD instances in long-tailed learning. We show that our method not only re-balances the training dataset, but also promotes the neural network to learn more separable representations. Besides, we also present a guideline about the open-set auxiliary datasets: the realism and sample size are more important than the diversity.
Extensive experiments demonstrate that our proposed method can enhance the performance of existing state-of-the-art methods, and also achieves strong OOD detection performance under class-imbalanced setting. Overall, our method enables to exploit OOD data for long-tail learning and we expect that our insights inspire future research to further explore the benefits of auxiliary datasets.

\section*{Acknowledgements}

This research is supported by MOE Tier-1 project RG13/19 (S). LF is supported by the National Natural Science Foundation of China (Grant No. 62106028) and CAAI-Huawei MindSpore Open Fund.

\bibliography{main}
\bibliographystyle{icml2022}

\newpage
\appendix
\onecolumn

\section{Proof of Theorem~\ref{thm:bayes_fix}}
\label{app:proofs_1}

Since $\dmix = \dt \cup \dout$, the underlying data distribution of $\dmix$ will be a linear combination of the training distribution $\pt(X, Y)$ and the OOD distribution $\pout(X, Y)$:
\begin{equation}
\pmix(\vx, y) = \frac{N}{M+N} \pt(\vx, y) + \frac{M}{M+N} \pout(\vx, y),
\end{equation}
where $N$ is the size of $\dt$ and $M$ is the size of $\dout$. 

By the virtue of Bayes’ theorem, we have
\begin{equation}
\begin{aligned}
& \pmix(\vx|y) \pmix(y) \\
=& \frac{N}{M+N} \pt(\vx|y) \pt(y) + \frac{M}{M+N} \pout(\vx|y) \pout(y) \\
=& \frac{N}{M+N} \pt(\vx|y) \pt(y) + \frac{M}{M+N} \pout(\vx) \pout(y) \\
=& \frac{N}{M+N} \pt(\vx|y) \pt(\vx, y) + \frac{1}{K} \cdot \frac{M}{M+N} \pout(\vx), \\
\end{aligned}
\end{equation}
where the second equality follows the fact that $\pmix(\vx|y)=\pmix(\vx)$ since the label $y$ is independent of the instance $x$ for the OOD data, and the third equality is simply the fact that $\pout(y)=1/K$.

Then, by taking the maximum of both sides, we have
\begin{equation}
\begin{aligned}
& \underset{y\in\cY}{\arg\max}~\pmix(\vx|y) \pmix(y) \\
=& \underset{y\in\cY}{\arg\max}~\left\{ \frac{N}{M+N} \pt(\vx|y) \pt(\vx, y) + \frac{M/K}{M+N} \pout(\vx) \right\} \\
=& \underset{y\in\cY}{\arg\max}~\frac{N}{M+N} \pt(\vx|y) \pt(\vx, y). \\
=& \underset{y\in\cY}{\arg\max}~\pt(\vx|y) \pt(\vx, y).
\end{aligned}
\end{equation}

Thus Theorem~\ref{thm:bayes_fix} is proved.
\qed

\section{Proof of Proposition~\ref{prop:complementary}}
\label{app:proofs_2}
By definition, $\sum^{K}_{i=1} \Gamma_i = 1$ naturally holds for any $\alpha$.

If $\alpha = \max_j (\beta_j)$, then $\Gamma_j = (\max_j (\beta_j) - \beta_j)/(K \cdot \max_j (\beta_j) - 1)$. In particular, for $k=\arg\max_i(\beta_i)$, we have $\Gamma_k=0$.

In the case of $\alpha \to \infty$, let us denote $f(\alpha) = \alpha - \beta_j$ and $g(\alpha) = K \cdot \alpha-1 $. Since $\lim_{\alpha \to \infty} f(\alpha) = \lim_{\alpha \to \infty} g(\alpha) = \infty$, $g^{\prime}(\alpha) = K \neq 0$, and $\lim_{\alpha \to \infty} f^{\prime}(\alpha)/g^{\prime}(\alpha) = 1/K$ exists, using L'Hôpital's rule, we have:
$$ \lim_{\alpha \to \infty} \Gamma_{j} = \lim_{\alpha \to \infty} \frac{f(\alpha)}{g(\alpha)} = \lim_{\alpha \to \infty} \frac{f^{\prime}(\alpha)}{g^{\prime}(\alpha)} = 1/K$$
Thus Proposition~\ref{prop:complementary} is proved.
\qed

\section{Algorithm Details}
\label{app:alg}

The details of Open-sampling are provided below:

\begin{algorithm}[ht]
\caption{Open-sampling}
\label{alg:open-sampling}
\begin{algorithmic}[1]
\REQUIRE Training dataset $\mathcal{D}_{\mathrm{train}}$. Open-set auxiliary dataset $\mathcal{D}^{(x)}_{\mathrm{out}}$;\\

\FOR{each iteration}
\STATE Sample a mini-batch of original training samples $\{(\boldsymbol{x}_i, y_i)\}^{n}_{i=0}$ from $\mathcal{D}_{\mathrm{train}}$;\\
\STATE Sample a mini-batch of open-set instances $\{\widetilde{\boldsymbol{x}}_i\}^{m}_{i=0}$ from $\mathcal{D}^{(x)}_{\mathrm{out}}$;\\
\STATE Generate random noisy label $\widetilde{y}_i \sim \Gamma $ for each open-set instance $\widetilde{\boldsymbol{x}}_i$;\\
\STATE Perform gradient descent on $f$ with $\mathcal{L}_{\mathrm{total}}$ from Equation (\ref{eq:final_loss});\\
\ENDFOR

\label{code:open_reg}
\end{algorithmic}
\end{algorithm}

\section{Datasets and empirical settings}
\label{sec:dataset}

\noindent\textbf{Long-Tailed CIFAR.} The original version of CIFAR-10 and CIFAR-100 contains 50,000 training images and 10,000 validation images of size $32 \times 32$ with 10 and 100 classes, respectively. To create their long-tailed version, we reduce the number of training examples per class according to an exponential function $n = n_j \mu^j$, where $j$ is the class index, $n_j$ is the original number of training images, and $\mu \in (0,1)$. Besides, the validation set and the test set are kept unchanged. The imbalance ratio of a dataset is defined as the number of training samples in the largest class divided by that of the smallest.

\noindent\textbf{CelebA-5.} CelebFaces Attributes (CelebA) dataset is a real-world long-tailed dataset. It is originally composed of 202,599 number of RGB face images with 40 binary attributes annotations per image. Note that CelebA is originally a multi-labeled dataset, we port it to a 5-way classification task by filtering only the samples with five non-overlapping labels about hair colors. We also subsampled the full dataset by 1/20 while maintaining the imbalance ratio as 10.7, following \cite{kim2020m2m}. In particular, We pick out 50 and 100 samples in each class for validation and testing. We denote the resulting dataset by CelebA-5. 

\noindent\textbf{Places-LT.} Places-LT is a long-tailed version of the large-scale scene classification dataset Places \cite{zhou2017places}. It consists of 184.5K images from 365 categories with class cardinality ranging from 5 to 4,980.

\noindent\textbf{Auxiliary datasets}. 300K Random Images \cite{hendrycks2019oe} is a cleaned and debiased dataset with 300K natural images. In this dataset, Images that belong to CIFAR classes from it, images that belong to Places or LSUN classes, and images with divisive metadata are removed so that $\mathcal{D}_{\mathrm{train}}$ and $\mathcal{D}_{\mathrm{out}}$ are disjoint.
We use the dataset as the open-set auxiliary dataset for experiments with CIFAR-10/100 and CelebA-5. For Experiments on Places-LT, we use the training set of Places-Extra69 \cite{zhou2017places} as the open-set auxiliary dataset.

\noindent\textbf{OOD test datasets}. Following OE \cite{hendrycks2019oe}, we comprehensively evaluate OOD detectors on artificial and real anomalous distributions, including: Gaussian, Rademacher, Blobs, Textures \cite{cimpoi2014describing}, SVHN \cite{netzer2011reading}, Places365 \cite{zhou2017places}, LSUN-Crop \cite{yu2015lsun}, LSUN-Resize \cite{yu2015lsun}, iSUN \cite{xu2015turkergaze}. For experiments on CIFAR-10, we also use CIFAR-100 as OOD test dataset. \textit{Gaussian} noises have each dimension i.i.d. sampled from an isotropic Gaussian distribution. \textit{Rademacher} noises are images where each dimension is $-1$ or $1$ with equal probability, so each dimension is sampled from a symmetric Rademacher distribution. \textit{Blobs} noises consist in algorithmically generated amorphous shapes with definite edges. 
\textit{Textures} \cite{cimpoi2014describing} is a dataset of describable textural images. 
\textit{SVHN} dataset \cite{netzer2011reading} contains 32 × 32 color images of house numbers. There are ten classes comprised of the digits 0-9.
\textit{Places365} \cite{zhou2017places} consists in images for scene recognition rather than object recognition.
\textit{LSUN} \cite{yu2015lsun} is another scene understanding dataset with fewer classes than Places365. Here we use \textit{LSUN-Crop} and \textit{LSUN-Resize} to denote the cropped and resized version of the LSUN dataset respectively.
\textit{iSUN} \cite{xu2015turkergaze} is a large-scale eye tracking dataset, selected from natural scene images of the SUN database \cite{xiao2010sun}. 

\section{Implementation details}
\label{app:settings}
For experiments on Long-Tailed CIFAR-10/100 \cite{krizhevsky2009learning} and CelebA-5 \cite{liu2015deep}, we perform training with ResNet-32 \cite{he2016deep} for 200 epochs, using SGD with a momentum of 0.9, and a weight decay of 0.0002. We set the initial learning rate as 0.1, then decay by 0.01 at the 160th epoch and again at the 180th epoch. For fair comparison, We also use linear warm-up learning rate schedule \cite{goyal2017accurate} for the first 5 epochs. For data augmentation in training, we use the commonly used version: 4 pixels are padded on each side, and a $32 \times 32$ crop is randomly sampled from the padded image or its horizontal flip. For experiments under the setting of Balanced Softmax \cite{Ren2020balms}, we use Nesterov SGD with momentum 0.9 and weight-decay 0.0005 for training. We use a total mini-batch size of 512 images on a single GPU. The learning rate increased from 0.05 to 0.1 in the first 800 iterations. Cosine scheduler \cite{loshchilov2016sgdr} is applied afterward, with a minimum learning rate of 0. Our augmentation follows Balanced Softmax and Equalization Loss \cite{tan2020equalization}. In testing, the image size is $32\times32$. In end-to-end training, the model is trained for 13K iterations. In decoupled training experiments, we fix the Softmax model, i.e., the instance-balanced baseline model obtained from the standard end-to-end training, as the feature extractor. And the classifier is trained for 2K iterations. 

For Places-LT, we choose ResNet-152 as the backbone network and pretrain it on the full ImageNet-2012 dataset, following Decoupled training \cite{kang2019decoupling}. We use SGD optimizer with momentum 0.9, batch size 512, cosine learning rate schedule \cite{loshchilov2016sgdr}. Similar to Decoupled training, we fine-tune the backbone with Instance-balanced sampling for representation learning and then re-train the classifier with our proposed algorithm with class-balanced sampling. Through the paper, we refer to decoupled training as training the last linear classifier on a fixed feature extractor obtained from instance-balanced training. For experiments with the SSP method, we use rotation prediction \cite{komodakis2018unsupervised} as self-supervised learning method, where an image is rotated by a random multiple of 90 degrees and a model is trained to predict the rotation degree as a 4-way classification problem. 

We conduct all the experiments on NVIDIA GeForce RTX 3090, and implement all methods with default parameters by PyTorch \cite{paszke2019pytorch}. All experiments are repeated five times with different seeds and we report the average test accuracy. We tune the hyperparameter $\eta$ on the validation set, then train the model on the full training set. The $\alpha$ is fixed as $(\max_j \beta_j + \min_j \beta_j)$ by default and we find this value performs well overall. For the $\eta$ in the training objective, the best value depends on the dataset, imbalance ratio, network architecture, and the integrated method.

For OOD detection task, we measure the following metrics: (1) the false positive rate (FPR95) of OOD examples when the true positive rate of in-distribution examples is at 95\%; (2) the area under the receiver operating characteristic curve (AUROC); and (3) the area under the precision-recall curve (AUPR).

\section{More empirical results}
\label{app:results}

\subsection{Ablation study}

To show the effect of the class-dependent weighting factor $w_j$ in Open-sampling, we conduct ablation study on long-tailed CIFAR-10 with imbalanced rate 100. As shown in Table \ref{tab:cifar10_ablation}, the class-dependent weighting factor generally improves the performance, when compared with the variant without the weight.

 \begin{table}[ht]
\centering
\caption{Results of ablation study on long-tailed CIFAR-10 for the class-dependent weighting factor $w_j$.}
\label{tab:cifar10_ablation}
\renewcommand\arraystretch{0.9}
\resizebox{0.47\textwidth}{!}{
\setlength{\tabcolsep}{5.3mm}{
\begin{tabular}{c|ccc}
\toprule
Imbalance Factor & 100 & 50 & 10 \\
\midrule
Standard & 71.61 & 77.30 & 86.74  \\
Ours w/o $w_j$ & 76.57 & 81.18 & 88.57  \\
Ours & \textbf{77.62} & \textbf{81.76} & \textbf{89.38} \\

\bottomrule
\end{tabular}
}
}
\end{table}

\subsection{Detailed results for OOD detection.}

Table \ref{tab:ood_detail} presents the detailed results of OOD detection performance on long-tailed CIFAR-10 with imbalanced rate 100 with various OOD test datasets. From the results, we show that our method can outperform OE \cite{hendrycks2019oe} in OOD detection tasks when training dataset is class-imbalanced.

\begin{table*}[!t]
\centering
\renewcommand\arraystretch{1}
\caption{ OOD detection performance comparison on long-tailed CIFAR-10 with imbalanced rate 100. $\uparrow$ indicates larger values are better and $\downarrow$ indicates smaller values are better.} 
\label{tab:ood_detail}
\setlength{\tabcolsep}{5mm}{
\begin{tabular}{ccccc}
\toprule
 OOD test dataset & Method & FPR95 $\downarrow$	& AUROC $\uparrow$		& AUPR $\uparrow$	\\
\midrule
\multirow{4}*{Gaussian}  & MSP&43.02 & 68.62 & 21.54 \\
 &OE&13.71  & 87.94 & 40.61 \\
& Ours&7.52  & 93.37 & 53.82 \\
& Ours $(\alpha=5)$ &1.20  & 99.55 & 95.72 \\
\midrule
\multirow{4}*{Rademacher} & MSP&38.33 & 81.44 & 32.97 \\
 &OE&13.88  & 87.17 & 39.50 \\
& Ours&7.93  & 92.91 & 52.39 \\
& Ours $(\alpha=5)$ &3.25  & 97.90 & 78.00 \\
\midrule
\multirow{4}*{Blob} & MSP&49.03 & 80.88 &  39.51 \\
 &OE&44.29  & 76.55 & 27.28 \\
& Ours&9.42  & 93.33  & 55.45 \\
& Ours $(\alpha=5)$ &33.65   & 91.50 &  66.51 \\
\midrule
\multirow{4}*{Textures} & MSP&66.53 & 72.19 & 29.27 \\
 &OE&38.00  & 84.92 & 38.14 \\
& Ours&23.29  & 92.53 & 57.99 \\
& Ours $(\alpha=5)$ &24.64  & 94.91 & 78.43 \\
\midrule
\multirow{4}*{SVHN} & MSP&63.74 & 69.11 & 24.06 \\
 &OE&43.46  & 84.17 & 37.06 \\
& Ours& 22.09  & 93.53 &  63.92 \\
& Ours $(\alpha=5)$ &28.89  & 92.12 & 67.60 \\
\midrule
\multirow{4}*{CIFAR-100} & MSP&70.50 & 70.24 & 28.27 \\
 &OE&61.15  & 77.88 & 32.77 \\
& Ours&55.32  & 84.09 & 45.81 \\
& Ours $(\alpha=5)$ &54.73  & 84.75 & 54.57 \\
\midrule
\multirow{4}*{LSUN-C} & MSP&40.68 & 85.52 & 50.07 \\
 &OE&22.60  & 86.80 & 39.06 \\
& Ours&14.73   & 96.70 & 80.81 \\
& Ours $(\alpha=5)$ &13.66  & 96.25 & 82.77 \\
\midrule
\multirow{4}*{LSUN-R} & MSP&59.27 & 76.86 & 36.88 \\
 &OE&17.91  & 87.36 & 39.86 \\
& Ours&15.29  & 93.87 & 60.49 \\
& Ours $(\alpha=5)$ &10.68  & 97.10 & 82.44  \\
\midrule
\multirow{4}*{iSUN} & MSP&61.86 & 74.66 & 32.68 \\
 &OE&20.02   & 87.44 & 40.37 \\
& Ours&15.33  & 93.66 & 58.85 \\
& Ours $(\alpha=5)$ &11.47  & 96.92 & 82.40 \\
\midrule
\multirow{4}*{Places365} & MSP& 67.98 & 72.47 & 31.82 \\
 &OE&48.77  & 81.24 & 33.90  \\
& Ours&35.91  & 89.98 & 54.23 \\
& Ours $(\alpha=5)$ &39.16  & 91.60 & 70.69 \\
\bottomrule
\end{tabular}
}
\vspace{-10pt}
\end{table*}

\end{document}